\documentclass[sigconf]{acmart}

\usepackage{booktabs} 

\usepackage{xcolor}
\usepackage{graphicx}
\usepackage{bm}
\usepackage{amsmath}
\usepackage{amssymb}
\usepackage{caption}
\usepackage{subcaption}
\usepackage{hyperref}
\usepackage{cleveref}
\usepackage{diagbox}
\usepackage{booktabs}
\usepackage{multirow,tabularx}
\usepackage{threeparttable}
\usepackage{xspace}
\usepackage{tabularx}
\usepackage[normalem]{ulem}

\crefname{chapter}{Chapter}{Chapters}
\crefname{section}{$\S$}{$\S$}
\crefname{subsection}{$\S$}{$\S$}
\crefname{subsubsection}{$\S$}{$\S$}
\crefname{figure}{Fig}{Figs}
\crefname{equation}{Eqn}{Eqns}
\crefname{table}{Table}{Tables}

\newcolumntype{Y}{>{\centering\arraybackslash}X}

\newif\ifdevelop\developfalse

\newif\ifcompact\compactfalse
\newif\ifnotcompact\notcompacttrue

\ifcompact
  \usepackage[skip=5pt]{caption}
  \usepackage[skip=3pt]{subcaption}
\else
  \usepackage{caption}
  \usepackage{subcaption}
\fi

\usepackage{setspace}
\ifcompact
  
\fi

\graphicspath{{figures/}}

\ifdevelop
  
\else
  
\fi

\ifdevelop
  \newcommand\devcomment[1]{\textcolor{red}{***{#1}***}}
\else
  \newcommand\devcomment[1]{}
\fi

\ifdevelop
  \newcommand\maybedeleted[1]{\textcolor{blue}{#1}}
\else
  \newcommand\maybedeleted[1]{}
\fi

\ifcompact
  \ifdevelop
    \newcommand\compactdel[1]{{\textcolor{red}{#1}}}
    \newcommand\compactsub[2]{{\textcolor{red}{#1}}{\textcolor{blue}{#2}}}
  \else
    \newcommand\compactdel[1]{}
    \newcommand\compactsub[2]{#2}
  \fi
  \newcommand\compactvspace[1]{\vspace{#1}}
\else
  \newcommand\compactdel[1]{#1}
  \newcommand\compactsub[2]{#1}
  \newcommand\compactvspace[1]{}
\fi

\newcommand{\disabled}[1]{}
\newcommand{\numberthis}{\stepcounter{equation}\tag{\theequation}}

\newcommand{\DONUT}{\textit{Donut}}
\newcommand{\IE}{\textit{i.e.}}
\newcommand{\EG}{\textit{e.g.}}
\newcommand{\WRT}{\textit{w.r.t.}}
\newcommand{\DATASETA}{$\mathcal{A}$}
\newcommand{\DATASETB}{$\mathcal{B}$}
\newcommand{\DATASETC}{$\mathcal{C}$}

\newcommand{\dd}{\mathrm{d}}
\newcommand{\vv}[1]{\bm{\mathrm{{#1}}}}
\newcommand{\E}{\operatorname{\mathbb{E}}}
\newcommand{\EE}[1]{\operatorname{\mathbb{E}}\left[{#1}\right]}
\newcommand{\EEE}[2]{\operatorname{\mathbb{E}}_{{#1}}\left[{#2}\right]}

\newcommand{\KLDD}[2]{\operatorname{KL}\left[{#1}\,\big\|\,{#2}\right]}
\newcommand{\abs}[1]{\left|#1\right|}

\newcommand{\Entropyy}[1]{\operatorname{H}\left[#1\right]}

\setcopyright{rightsretained}

\copyrightyear{2018}
\acmYear{2018}
\setcopyright{iw3c2w3}
\acmConference[WWW 2018]{The 2018 Web Conference}{April 23--27, 2018}{Lyon, France}
\acmBooktitle{WWW 2018: The 2018 Web Conference, April 23--27, 2018, Lyon, France}
\acmPrice{}
\acmDOI{https://doi.org/10.1145/3178876.3185996}
\acmISBN{978-1-4503-5639-8}

\ifcompact
\fi

\begin{document}
\title{Unsupervised Anomaly Detection via Variational Auto-Encoder \\for Seasonal KPIs in Web Applications}

\author{Haowen Xu, Wenxiao Chen, Nengwen Zhao, Zeyan Li, Jiahao Bu, Zhihan Li, Ying Liu, Youjian Zhao, Dan Pei}
\authornote{Dan Pei is the corresponding author.}
\affiliation{Tsinghua University}
\author{Yang Feng, Jie Chen, Zhaogang Wang, Honglin Qiao}
\affiliation{Alibaba Group}

\renewcommand{\shortauthors}{Haowen Xu et al.}

\begin{abstract}

To ensure undisrupted business, large Internet companies need to closely monitor various KPIs (\EG, Page Views, number of online users, and number of orders) of its Web applications,  to accurately detect anomalies and trigger timely troubleshooting/mitigation. However, anomaly detection for these seasonal KPIs with various patterns and data quality has been a great challenge, especially without labels. 
In this paper, we proposed \DONUT{}, an unsupervised anomaly detection algorithm based on VAE.
Thanks to  a few of our key techniques,
\DONUT{}\footnote{An implementation of \DONUT{} is published at \url{https://github.com/korepwx/donut}} greatly outperforms a state-of-arts supervised ensemble approach and a baseline VAE approach, and its best F-scores range from 0.75 to 0.9 for the studied KPIs from a top global Internet company.
%
We come up with a novel KDE interpretation of reconstruction for \DONUT{}, making it the first VAE-based anomaly detection algorithm with solid theoretical explanation.

\end{abstract}


%

\ifnotcompact

\fi

\maketitle

\begin{abstract}

To ensure undisrupted business, large Internet companies need to closely monitor various KPIs (\EG, Page Views, number of online users, and number of orders) of its Web applications,  to accurately detect anomalies and trigger timely troubleshooting/mitigation. However, anomaly detection for these seasonal KPIs with various patterns and data quality has been a great challenge, especially without labels. 
In this paper, we proposed \DONUT{}, an unsupervised anomaly detection algorithm based on VAE.
Thanks to  a few of our key techniques,
\DONUT{}\footnote{An implementation of \DONUT{} is published at \url{https://github.com/korepwx/donut}} greatly outperforms a state-of-arts supervised ensemble approach and a baseline VAE approach, and its best F-scores range from 0.75 to 0.9 for the studied KPIs from a top global Internet company.
%
We come up with a novel KDE interpretation of reconstruction for \DONUT{}, making it the first VAE-based anomaly detection algorithm with solid theoretical explanation.

\end{abstract}


\section{Introduction}\label{sec:introduction}

To ensure undisrupted business, large Internet companies need to closely monitor various KPIs (key performance indicators) of its Web applications,  to accurately detect anomalies and trigger timely troubleshooting/mitigation. KPIs are time series data, measuring metrics such as  Page Views, number of online users, and number of orders. Among all KPIs, the most ones are business-related KPIs (the focus of this paper), which are heavily influenced by user behavior and schedule, thus roughly have seasonal patterns occurring at regular intervals (\EG, daily and/or weekly).  However, anomaly detection for these seasonal KPIs with various patterns and data quality has been a great challenge, especially without labels.

A rich body of literature exist on detecting KPI anomalies~\cite{ad-survey,historical-avg,MA,arma,arima,kalman,holt-winters,TSD,svd,wavelet,majority_vote,normalization_schema,opprentice,egads,one-class-svm1,deep-learning-svm,GMM, kde,vae-ad,vi-storn}.
As discussed in~\cref{sec:previous-work}, existing anomaly detection algorithms suffer from the hassle of algorithm picking/parameter tuning, heavy reliance on labels, unsatisfying performance, and/or lack of theoretical foundations.


In this paper, we propose \DONUT{}, an unsupervised anomaly detection algorithm based on Variational Auto-Encoder (a representative deep generative model) with solid theoretical explanation, and this algorithm can work when there are no labels at all, and can take advantage of the occasional labels when available.

The contributions of this paper can be summarized as follows.


\begin{itemize}

\item The three techniques in \DONUT{}, Modified ELBO and Missing Data Injection for training, and MCMC Imputation for detection, enable it to greatly outperform state-of-art supervised and VAE-based anomaly detection algorithms.  The best F-scores of unsupervised \DONUT{} range from 0.75 to 0.9 for the studied KPIs from a top global Internet company.

\item For the first time in the literature, we discover that adopting VAE (or generative models in general) for anomaly detection requires training on both normal data \textit{and abnormal data}, contrary to common intuition.

\item We propose a novel KDE interpretation in z-space for \DONUT{}, making it the first VAE-based anomaly detection algorithm with solid theoretical explanation unlike~\cite{vae-ad,vi-storn}. This interpretation may benefit the design of other deep generative models in anomaly detection. We discover a \textit{time gradient effect} in latent z-space, which nicely explain \DONUT{}'s excellent performance for detecting anomalies in seasonal KPIs.
\end{itemize}




\section{Background and Problem}
\label{sec:background}

\subsection{Context and Anomaly Detection in General}
\label{sec:focus-of-this-paper}

\begin{figure}
	\centering
	\includegraphics[width=\columnwidth]{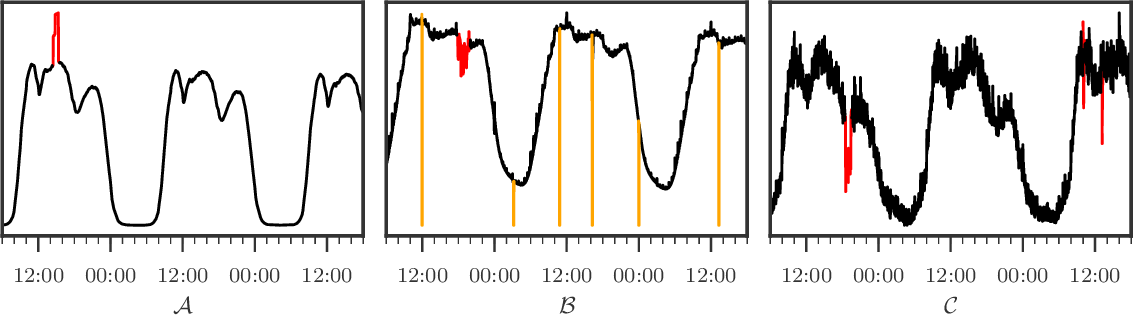}
	\caption{
		2.5-day-long fragments of the seasonal KPI datasets in our paper, with anomalies in red color and missing points (filled with zeros) in orange. Within each dataset, there are variations for the same time slot in different days.
	}
	\label{fig:kpi}
\end{figure}

In this paper, we focus on business-related KPIs. These time series  are heavily influenced by user behavior and schedule, thus roughly have \textit{seasonal} patterns occurring at regular intervals (\EG, daily and/or weekly). On the other hand, the \textit{shapes} of the KPI curves at each repetitive cycle are \textit{not}  exactly the same, since user behavior can vary across days. We hereby name the KPIs we study \textbf{``seasonal KPIs with local variations''}. Examples of such KPIs are shown in \cref{fig:kpi}. Another type of local variation is the increasing trend over days, as can be identified by Holt-Winters~\cite{holt-winters} and Time Series Decomposition~\cite{TSD}.
An anomaly detection algorithm may not work well unless these local variations are properly handled.

In addition to the seasonal patterns and local variations of the KPI \textit{shapes}, there are also noises on these KPIs, which we assume to be independent, zero-mean Gaussian at every point.
The exact values of the Gaussian noises are meaningless, thus we only focus on the statistics of these noises, \IE, the variances of the noises.

We can now formalize the ``normal patterns'' of seasonal KPIs as a combination of two components: (1) the seasonal patterns with local variations, and (2) the statistics of the Gaussian noises.

We use ``anomalies'' to denote the recorded points which do not follow normal patterns (\EG, sudden spikes and dips) , while using ``abnormal'' to denote both anomalies and missing points. See \cref{fig:kpi} for examples of both anomalies and missing points.
 Because the KPIs are monitored periodically (\EG, every minute), missing points are recorded as ``null'' (when the monitoring system does not receive the data) and thus are straightforward to identify.   We thus focus on detecting anomalies for the KPIs.

Because operators need to deal with the anomalies for troubleshooting/mitigation, some of the anomalies are anecdotally labeled. Note that such occasional labels' coverage of anomalies are far from what's needed for typical supervised learning algorithms.

\textbf{Anomaly detection} on KPIs can be formulated as follows: for any time $t$, given historical observations $x_{t-T+1}, \dots, x_t$, determine whether an anomaly occurs (denoted by $y_t=1$).
An anomaly detection algorithm typically computes a real-valued score indicating the certainty of having $y_t=1$, \EG, $p(y_t=1|x_{t-T+1},\dots,x_t)$, instead of directly computing $y_t$.
Human operators can then affect whether to declare an anomaly by choosing a threshold, where a data point with a score exceeding this threshold indicates an anomaly.


\compactvspace{-1.5em}
\subsection{Previous Work}
\label{sec:previous-work}

\textbf{Traditional statistical models.} Over the years, quite a few anomaly detectors based on traditional statistical models (\textit{e.g.}, \cite{historical-avg,MA,arma,arima,kalman,holt-winters,TSD,svd,wavelet}, mostly time series models) have been proposed to compute anomaly scores. Because these algorithms typically have simple assumptions for applicable KPIs,  expert's efforts need to be involved to pick a suitable detector for a given KPI, and then fine-tune the detector's parameters based on the training data.  Simple ensemble of these detectors, such as majority vote~\cite{majority_vote} and normalization~\cite{normalization_schema}, do not help much either according to~\cite{opprentice}. As a result, these detectors see only limited use in the practice.

\textbf{Supervised ensemble approaches.}
To circumvent the hassle of algorithm/parameter tuning for traditional statistical anomaly detectors, supervised ensemble approaches, EGADS~\cite{egads} and Opprentice~\cite{opprentice}, were proposed. They train anomaly classifiers using the user feedbacks as labels and using anomaly scores output by traditional detectors as features. Both EGADS and Opprentice showed promising results, but they heavily rely on good labels (much more than the anecdotal labels accumulated in our context), which is generally not feasible in large scale applications. Furthermore, running multiple traditional detectors to extract features during detection introduces lots of computational cost, which is a practical concern. 

\textbf{Unsupervised approaches and deep generative models.}
Recently, there is a rising trend of adopting unsupervised machine learning algorithms for anomaly detection, \EG, one-class SVM~\cite{one-class-svm1,deep-learning-svm}, clustering based methods~\cite{cluster} like K-Means~\cite{k-means} and GMM~\cite{GMM}, KDE~\cite{kde}, and VAE~\cite{vae-ad} and VRNN~\cite{vi-storn}. The philosophy is to focus on normal patterns instead of anomalies: since the KPIs are typically composed mostly of normal data, models can be readily trained even without labels.
Roughly speaking, they all first recognize ``normal'' regions in the original or some latent feature space, and  then compute the anomaly score by measuring ``how far'' an observation is from the normal regions. 

Along this direction, we are interested in deep generative models for the following reasons. First,  learning normal patterns can be seen as learning the distribution of training data, 
which is a topic of generative models. Second, great advances have been achieved recently to train generative models with deep learning techniques, \EG, GAN~\cite{gan} and deep Bayesian network~\cite{prml,deep-bayes}.
The latter is family of deep generative models, which adopts the graphical~\cite{graph} model framework and variational techniques~\cite{variational}, with the VAE~\cite{kingma_auto-encoding_2014,rezende_stochastic_2014} as a representative work.
Third, despite deep generative model's great promise in anomaly detection, existing VAE-based anomaly detection method~\cite{vae-ad} was not designed for KPIs (time series), and does not perform well in our settings (see \cref{sec:experiments}), and there is no theoretical foundation to back up its designs of  deep generative models for anomaly detection (see \cref{sec:analysis}).  
Fourth, simply adopting the more complex models~\cite{vi-storn} based on VRNN shows long training time and poor performance in our experiments.
Fifth,  \cite{vae-ad} assumes training only on clean data, which is infeasible in our context,  while \cite{vi-storn} does not discuss this problem. 

\compactvspace{-.5em}
\subsection{Problem Statement}
\label{sec:problem}
In summary, existing anomaly detection algorithms suffer from the hassle of algorithm picking/parameter tuning, heavy reliance on labels, unsatisfying performance, and/or lack of theoretical foundations. Existing approaches are either unsupervised, or supervised but depending heavily on labels. However, in our context, labels are occasionally available although far from complete, which should be somehow taken advantage of. 

The problem statement of this paper is as follows. \textbf{We aim at an unsupervised anomaly detection algorithm based on deep generative models with solid theoretical explanation, and this algorithm can take advantage of the occasionally available labels.} Because VAE is a basic building block of deep Bayesian network, we chose to start our work with VAE. 


\compactvspace{-.5em}
\subsection{Background of Variational Auto-Encoder}
\label{sec:variational-auto-encoder}

Deep Bayesian networks use neural networks to express the relationships between variables, such that they are no longer restricted to simple distribution families, thus can be easily applied to complicated data.
Variational inference techniques~\cite{deep_learning_book} are often adopted in training and prediction, which are efficient methods to solve posteriors of the distributions derived by neural networks.

\begin{figure}
	\centering
	\ifcompact
	  \includegraphics[height=0.65in]{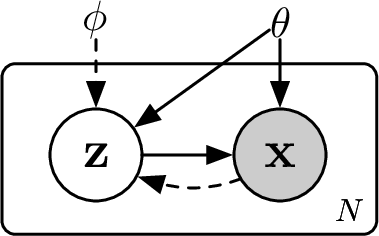}
	\else
	  \includegraphics[height=1.2in]{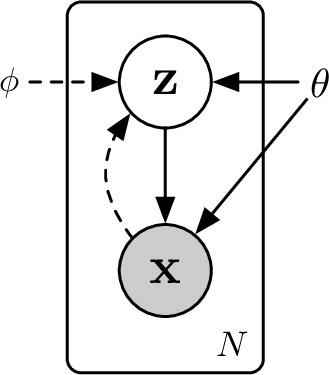}
	\fi
	\caption{Architecture of VAE.  The prior of $\vv{z}$ is regarded as part of the generative model (solid lines), thus the whole generative model is denoted as $p_{\theta}(\vv{x},\vv{z}) = p_{\theta}(\vv{x}|\vv{z})\,p_{\theta}(\vv{z})$. The approximated posterior (dashed lines) is denoted as $q_{\phi}(\vv{z}|\vv{x})$.}
	\label{fig:vae}
\end{figure}

VAE is a deep Bayesian network.
It models the relationship between two random variables, latent variable $\vv{z}$ and visible variable $\vv{x}$.
A prior is chosen for $\vv{z}$, which is usually multivariate unit Gaussian $\mathcal{N}(\vv{0},\vv{I})$.
After that, $\vv{x}$ is sampled from $p_{\theta}(\vv{x}|\vv{z})$, which is derived from a neural network with parameter $\theta$.
The exact form of $p_{\theta}(\vv{x}|\vv{z})$ is chosen according to the demand of task.
The true posterior $p_{\theta}(\vv{z}|\vv{x})$ is intractable by analytic methods, but is necessary for training and often useful in prediction, thus the variational inference techniques are used to fit another neural network as the approximation posterior $q_{\phi}(\vv{z}|\vv{x})$.
This posterior is usually assumed to be $\mathcal{N}(\vv{\mu}_{\phi}(\vv{x}),\vv{\sigma}^2_{\phi}(\vv{x}))$, where $\vv{\mu}_{\phi}(\vv{x})$ and $\vv{\sigma}_{\phi}(\vv{x})$ are derived by neural networks.
The architecture of VAE is shown as \cref{fig:vae}.

SGVB~\cite{kingma_auto-encoding_2014,rezende_stochastic_2014} is a variational inference algorithm that is often used along with VAE, where the approximated posterior and the generative model are jointly trained by maximizing the evidence lower bound (ELBO, \cref{eqn:vae-elbo}).
We did not adopt more advanced variational inference algorithms, since SGVB already works.
\begin{align*}
	\log p_{\theta}(\vv{\vv{x}}) &\geq \log p_{\theta}(\vv{x}) - \KLDD{q_{\phi}(\vv{z}|\vv{x})}{p_{\theta}(\vv{z}|\vv{x})} \\
		&= \mathcal{L}(\vv{x}) \numberthis\label{eqn:vae-elbo} \\
		&= \EEE{q_{\phi}(\vv{z}|\vv{x})}{\log p_{\theta}(\vv{x}) + \log p_{\theta}(\vv{z}|\vv{x}) - \log q_{\phi}(\vv{z}|\vv{x})} \\
		&= \EEE{q_{\phi}(\vv{z}|\vv{x})}{\log p_{\theta}(\vv{x},\vv{z}) - \log q_{\phi}(\vv{z}|\vv{x})} \\
		&= \EEE{q_{\phi}(\vv{z}|\vv{x})}{\log p_{\theta}(\vv{x}|\vv{z}) + \log p_{\theta}(\vv{z}) - \log q_{\phi}(\vv{z}|\vv{x})}
\end{align*}
Monte Carlo integration~\cite{geweke1989bayesian} is often adopted to approximate the expectation in \cref{eqn:vae-elbo}, as \cref{eqn:monte-carlo-integration}, where $\vv{z}^{(l)}, l=1 \dots L$ are samples from $q_{\phi}(\vv{z}|\vv{x})$.
We stick to this method throughout this paper.
\begin{equation}
	\EEE{q_{\phi}(\vv{z}|\vv{x})}{f(\vv{z})}
		\approx \frac{1}{L} \sum_{l=1}^L f(\vv{z}^{(l)})
	\label{eqn:monte-carlo-integration}
\end{equation}



\section{Architecture}
\label{sec:architecture}

\begin{figure}
	\centering
	\includegraphics[width=\columnwidth]{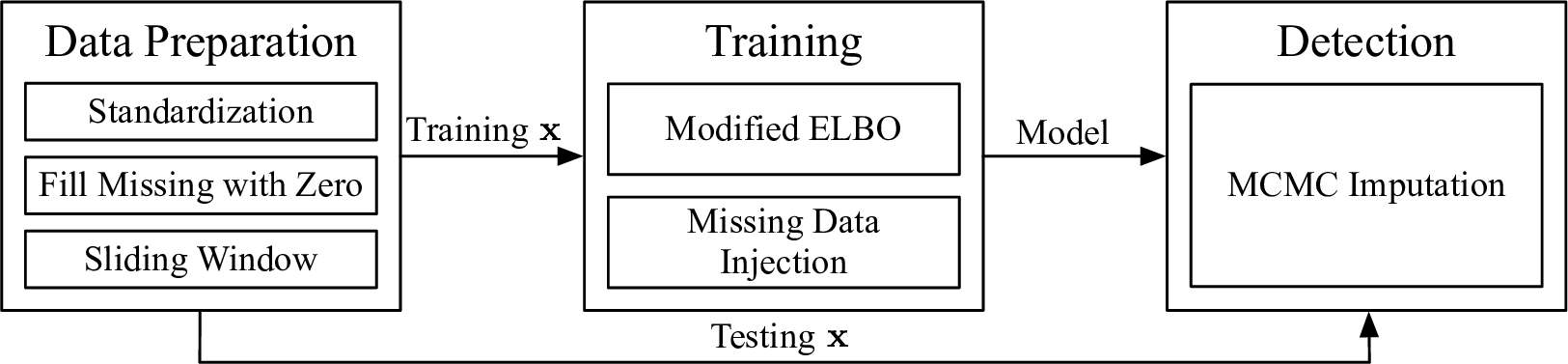}
	\caption{
		Overall architecture of \DONUT{}.
	}\label{fig:architecture}
\end{figure}

The overall architecture of our algorithm \DONUT{} is illustrated as \cref{fig:architecture}. The three key techniques are \textit{Modified ELBO} and \textit{Missing Data Injection} during training, and \textit{MCMC Imputation} in detection.

\compactvspace{-.6em}
\subsection{Network Structure}
\label{sec:network-structure}

As aforementioned in ~\cref{sec:focus-of-this-paper}, the KPIs studied in this paper are assumed to be time sequences with Gaussian noises.
However, VAE is not a sequential model, thus we apply sliding windows~\cite{sejnowski1987parallel} of length $W$ over the KPIs: for each point $x_t$, we use $x_{t-W+1}, \dots, x_t$ as the $\vv{x}$ vector of VAE.
This sliding window was first adopted because of its simplicity, but it turns out to actually bring an important and beneficial consequence, which will be discussed in \cref{sec:kde-interpretation}.

\begin{figure}
	\begin{subfigure}[t]{0.495\columnwidth}
		\centering
		\ifcompact
		  \includegraphics[height=1.5in]{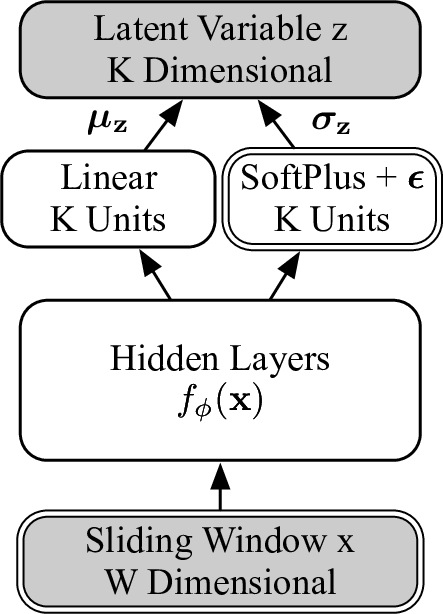}
		\else
		  \includegraphics[height=1.8in]{variational-net}
		\fi
		\caption{Variational net $q_{\phi}(\vv{z}|\vv{x})$}\label{fig:variational-net}
	\end{subfigure}\hfill
	\begin{subfigure}[t]{0.495\columnwidth}
		\centering
		\ifcompact
		  \includegraphics[height=1.5in]{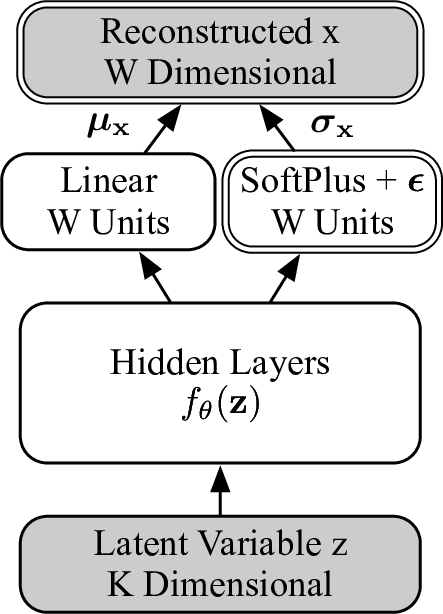}
		\else
		  \includegraphics[height=1.8in]{generative-net}
		\fi
		\caption{Generative net $p_{\theta}(\vv{x}|\vv{z})$}\label{fig:generative-net}
	\end{subfigure}

	\caption{
		Network structure of \DONUT{}.
		Gray nodes are random variables, and white nodes are layers.
		The double lines highlight our special designs upon a general VAE.
	}
	\label{fig:vae-network-structure}
\end{figure}

The overall network structure of \DONUT{} is illustrated in \cref{fig:vae-network-structure}, where the components with double-lined outlines (\EG, Sliding Window x, W Dimensional at bottom left) are our new designs and the remaining components are from standard VAEs.
The prior $p_{\theta}(\vv{z})$ is chosen to be $\mathcal{N}(\vv{0},\vv{I})$.
Both $\vv{x}$ and $\vv{z}$ posterior are chosen to be diagonal Gaussian: $p_{\theta}(\vv{x}|\vv{z}) = \mathcal{N}(\vv{\mu_x},\vv{\sigma_x}^2\vv{I})$, and $q_{\phi}(\vv{z}|\vv{x}) = \mathcal{N}(\vv{\mu_z},\vv{\sigma_z}^2\vv{I})$, where $\vv{\mu_x}$, $\vv{\mu_z}$ and $\vv{\sigma_x}$, $\vv{\sigma_z}$ are the means and standard deviations of each independent Gaussian component.
$\vv{z}$ is chosen to be $K$ dimensional.
Hidden features are extracted from $\vv{x}$ and $\vv{z}$, by separated hidden layers $f_{\phi}(\vv{x})$ and $f_{\theta}(\vv{z})$.
Gaussian parameters of $\vv{x}$ and $\vv{z}$ are then derived from the hidden features.
The means are derived from linear layers: $\vv{\mu_x} = \vv{W}^{\top}_{\vv{\mu_x}}f_{\theta}(\vv{z})+\vv{b_{\mu_x}}$ and $\vv{\mu_z} = \vv{W}^{\top}_{\vv{\mu_z}}f_{\phi}(\vv{x})+\vv{b_{\mu_z}}$.
The standard deviations are derived from soft-plus layers, plus a non-negative small number $\vv{\epsilon}$: $\vv{\sigma_x} = \operatorname{SoftPlus}[\vv{W}^{\top}_{\vv{\sigma_x}}f_{\theta}(\vv{z})+\vv{b_{\sigma_x}}] + \vv{\epsilon}$ and $\vv{\sigma_z} = \operatorname{SoftPlus}[\vv{W}^{\top}_{\vv{\sigma_z}}f_{\phi}(\vv{x})+\vv{b_{\sigma_z}}] + \vv{\epsilon}$, where $\operatorname{SoftPlus}[a] = \log [\exp (a) + 1]$.
All the $\vv{W}$-s and $\vv{b}$-s presented here are parameters of corresponding layers.
Note when scalar function $f(x)$ is applied on vector $\vv{x}$, it means to apply on every component.

We choose to derive $\vv{\sigma_x}$ and $\vv{\sigma_z}$ in such a way, instead of deriving $\log \vv{\sigma_x}$ and $\log \vv{\sigma_z}$ using linear layers as others do, for the following reason. The local variations in the KPIs of our interest are so small that $\vv{\sigma_x}$ and $\vv{\sigma_z}$ would probably get extremely close to zero, making $\log \vv{\sigma_x}$ and $\log \vv{\sigma_z}$ unbounded. This would cause severe numerical problems when computing the likelihoods of Gaussian variables.
We thus use the soft-plus and the $\vv{\epsilon}$ trick to prevent such problems.

We intentionally choose fully-connected layers as the structure of the hidden layers, making  the overall architecture fairly simple. This is because our objective is to develop an VAE based anomaly detection algorithm with solid theoretical explanation, and a simple network structure would definitely make it easier to analyze the internal behavior in the perplexing ``variational auto-encoder''.


\compactvspace{-.5em}
\subsection{Training}
\label{sec:training}

Training is straightforward by optimizing the ELBO (\cref{eqn:vae-elbo}) with SGVB~\cite{kingma_auto-encoding_2014} algorithm.
Since it is reported by \cite{kingma_auto-encoding_2014} that one sample is already sufficient for computing the ELBO when training VAE with the SGVB algorithm, we let sampling number $L=1$ during training.
We also apply the re-parameterization trick as required by SGVB: instead of sampling $\vv{z} \sim \mathcal{N}(\vv{\mu_z}, \vv{\sigma_z}^2 \vv{I})$, a dedicated random variable $\vv{\xi} \sim \mathcal{N}(\vv{0},\vv{I})$ is sampled, such that we can rewrite $\vv{z}$ as $\vv{z}(\vv{\xi}) = \vv{\mu_z} + \vv{\xi} \cdot \vv{\sigma_z}$.
Sampling on $\vv{\xi}$ is independent with the parameters $\phi$, which allows us to apply stochastic gradient descent as if VAE is an ordinary neural network.
The windows of $\vv{x}$ are randomly shuffled before every epoch, which is beneficial for stochastic gradient descent.
A sufficiently large number of $\vv{x}$ are taken in every mini-batch, which is critical for stabilizing the training, since sampling introduces extra randomness.

As discussed in \cref{sec:previous-work}, the VAE based anomaly detection works by learning normal patterns, thus we need to avoid learning abnormal patterns whenever possible. Note that the ``anomalies'' in training are labeled anomalies, and there can be no labels for a given KPI, in which case the anomaly detection becomes an unsupervised one.

One might be tempted to replace labeled anomalies (if any)
 and missing points (known) in training data with synthetically generated values. Some previous work has proposed methods to impute missing data, \EG, \cite{missing}, but it is hard to produce data that follow the ``normal patterns'' well enough. More importantly, training a generative model with data generated by another algorithm is quite absurd, since one major application of generative models is exactly to generate data.
Using data imputed by any algorithm weaker than VAE would potentially downgrade the performance.
Thus we do not adopt missing data imputation before training VAE, instead we choose to simply fill the missing points as zeros (in the \textit{Data Preparation} step in \cref{fig:architecture}), and then modify the ELBO to exclude the contribution of anomalies and missing points (shown as Modified ELBO (\textbf{M-ELBO} for short hereafter) in the \textit{Training} step in \cref{fig:architecture}).

More specifically, we modify the standard ELBO in \cref{eqn:vae-elbo} to our version \cref{eqn:vae-elbo-modified}.
$\alpha_w$ is defined as an indicator, where $a_w = 1$ indicates $x_w$ being not anomaly or missing, and $a_w = 0$ otherwise.
$\beta$ is defined as $(\sum_{w=1}^W \alpha_w) / W$. Note that  \cref{eqn:vae-elbo-modified} still holds when there is no labeled anomalies in the training data.
The contribution of $p_{\theta}(x_w|\vv{z})$ from labeled anomalies and missing points are directly excluded by $\alpha_w$, while the scaling factor $\beta$ shrinks the contribution of $p_{\theta}(\vv{z})$ according to the ratio of normal points in $\vv{x}$.
This modification trains \DONUT{} to correctly reconstruct the normal points within $\vv{x}$, even if some points in $\vv{x}$ are abnormal.
We do not shrink $q_{\phi}(\vv{z}|\vv{x})$, because of the following two considerations.
Unlike $p_{\theta}(\vv{z})$, which is part of the generative network (\IE, model of the ``normal patterns''), $q_{\phi}(\vv{z}|\vv{x})$ just describes the mapping from $\vv{x}$ to $\vv{z}$, without considering``normal patterns''.
Thus, discounting the contribution of $q_{\phi}(\vv{z}|\vv{x})$ seems not necessary.
Another reason is that $\E_{q_{\phi}(\vv{z}|\vv{x})}[-\log q_{\phi}(\vv{z}|\vv{x})]$ is exactly the entropy of $q_{\phi}(\vv{z}|\vv{x})$.
This entropy term actually has some other roles in training (which will be discussed in \cref{sec:clustering-effect-cause}), thus might be better kept untouched.
\compactvspace{-.4em}
\begin{equation}
\small{
	\widetilde{\mathcal{L}}(\vv{x})
		= \E_{q_{\phi}(\vv{z}|\vv{x})}\bigg[\sum_{w=1}^W \alpha_w \log p_{\theta}(x_w|\vv{z}) + \beta \log p_{\theta}(\vv{z}) - \log q_{\phi}(\vv{z}|\vv{x})\bigg]
	\label{eqn:vae-elbo-modified}
}
\end{equation}
Besides \cref{eqn:vae-elbo-modified}, another way to deal with anomalies and missing points is to exclude all windows containing these points from training data.
This approach turns out to be inferior to M-ELBO.
We will demonstrate the performance of both approaches in \cref{sec:trick-effects}.

Furthermore, we also introduce missing data injection in training: we randomly set $\lambda$ ratio of normal points to be zero, as if they are missing points.
With more missing points, \DONUT{} is trained more often to reconstruct normal points when given abnormal $\vv{x}$, thus the effect of M-ELBO is amplified.
This injection is done before every epoch, and the points are recovered once the epoch is finished. This missing data injection is shown in  the \textit{Training} step in \cref{fig:architecture}.


\compactvspace{-.5em}
\subsection{Detection}
\label{sec:detection}

Unlike discriminative models which are designed for just one purpose (\EG, a classifier is designed for just computing the classification probability $p(y|x)$), generative models like VAE can derive various outputs.
In the scope of anomaly detection, the likelihood of observation window $\vv{x}$, \IE, $p_{\theta}(\vv{x})$ in VAE, is an important output, since we want to see how well a given $\vv{x}$ follows the normal patterns.
Monte Carlo methods can be adopted to compute the probability density of $\vv{x}$, by $p_{\theta}(\vv{x}) = \EEE{p_{\theta}(\vv{z})}{p_{\theta}(\vv{x}|\vv{z})}$.
Despite the theoretically nice interpretation, sampling on the prior actually does not work well enough in practice, as will be shown in \cref{sec:experiments}.

Instead of sampling on the prior, one may seek to derive useful outputs with the variational posterior $q_{\phi}(\vv{z}|\vv{x})$.
One choice is to compute $\EEE{q_{\phi}(\vv{z}|\vv{x})}{p_{\theta}(\vv{x}|\vv{z})}$.
Although similar to $p_{\theta}(\vv{x})$, it is actually not a well-defined probability density.
Another choice is to compute $\EEE{q_{\phi}(\vv{z}|\vv{x})}{\log p_{\theta}(\vv{x}|\vv{z})}$, which is adopted in \cite{vae-ad}, named as ``reconstruction probability''.
These two choices are very similar.
Since only the ordering rather than the exact values of anomaly scores are concerned in anomaly detection, we follow \cite{vae-ad} and use the latter one. As an alternative, the ELBO (\cref{eqn:vae-elbo}) may also be used for approximating $\log p_{\theta}(\vv{x})$, as in \cite{vi-storn}.
However, the extra term $\EEE{q_{\phi}(\vv{z}|\vv{x})}{\log p_{\theta}(\vv{z}) - \log q_{\phi}(\vv{z}|\vv{x})}$ in ELBO makes its internal mechanism hard to understand.
Since the experiments in \cite{vi-storn} does not support this alternative's superiority, we choose not to use it.

During detection, the anomalies and missing points in a testing window $\vv{x}$ can bring bias to the mapped $\vv{z}$, and further make the reconstruction probability inaccurate, which would be discussed in \cref{sec:find-better-posterior}.
Since the missing points are always known (as ``null''), we have the chance to eliminate the biases introduced by missing points.
We choose to adopt the MCMC-based missing data imputation technique with the trained VAE, which is proposed by \cite{rezende_stochastic_2014}.
Meanwhile, we do not know the exact positions of anomalies before detection, thus MCMC cannot be adopted on anomalies.

\begin{figure}
	\centering
	\includegraphics[width=\columnwidth]{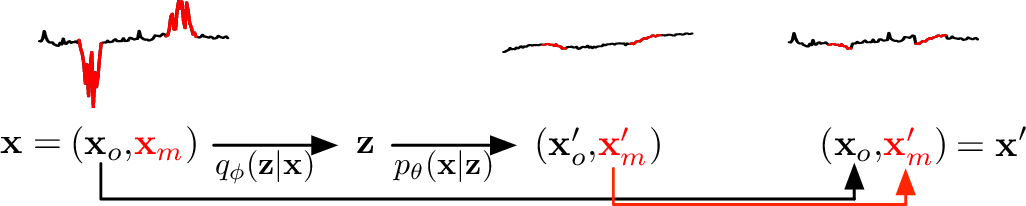}
	\caption{
		Illustration of one iteration in MCMC.
		$\vv{x}$ is decomposed as $(\vv{x}_o,\vv{x}_m)$, then $\vv{x}_o$ is fixed and $\vv{x}_m$ is replaced by $\vv{x}'_m$ from the reconstruction sample, in order to get the new $\vv{x}'$.
	}\label{fig:mcmc-illustration}
\end{figure}

More specifically, the testing $\vv{x}$ is divided into observed and missing parts, \IE, $(\vv{x}_o, \vv{x}_m)$.
A $\vv{z}$ sample is obtained from $q_{\phi}(\vv{z}|\vv{x}_o,\vv{x}_m)$, then a reconstruction sample $(\vv{x}'_o, \vv{x}'_m)$ is obtained from $p_{\theta}(\vv{x}_o,\vv{x}_m|\vv{z})$.
$(\vv{x}_o, \vv{x}_m)$ is then replaced by $(\vv{x}_o, \vv{x}'_m)$, \IE, the observed points are fixed and the missing points are set to new values.
This process is iterated for $M$ times, then the final $(\vv{x}_o, \vv{x}'_m)$ is used for computing the reconstruction probability.
The intermediate $\vv{x}'_m$ will keep getting closer to normal values during the whole procedure.
Given sufficiently large $M$, the biases can be reduced, and we can get a more accurate reconstruction probability.
The MCMC method is illustrated in \cref{fig:mcmc-illustration} and is shown in  the \textit{Detection} step in \cref{fig:architecture}.

After MCMC, we take $L$ samples of $\vv{z}$ to compute the reconstruction probability by Monte Carlo integration.
It is worth mentioning that, although we may compute the reconstruction probability for each point in every window of $\vv{x}$, we only use the score for the last point (\IE, $x_t$ in $x_{t-T+1},\dots,x_t$), since we want to respond to anomalies as soon as possible during the detection.
We will still use vector notations in later texts, corresponding to the architecture of VAE.
While it is possible to improve the detection performance by delaying the decision and considering more scores for the same point at different times, we leave it as a future work.



\section{Evaluation}
\label{sec:experiments}

\subsection{Datasets}

\compactdel{
\begin{figure}
	\centering
	\includegraphics[width=\columnwidth]{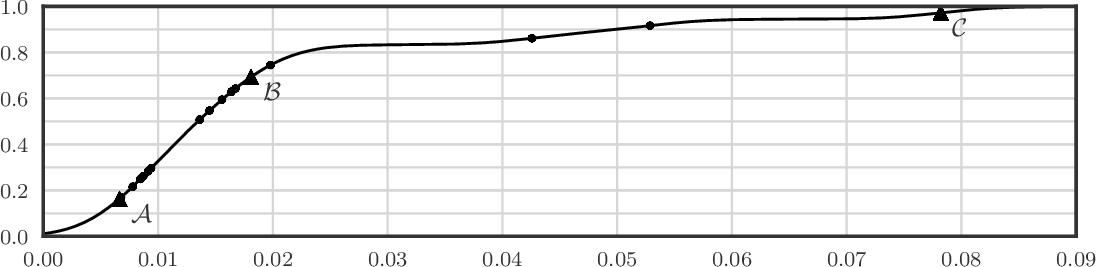}
	\caption{\compactdel{
		CDF of $\frac{1}{N} \sum_{t=2}^N \abs{x_t - x_{t-1}}$ of the datasets, which is plotted by placing Gaussian kernels at each dataset.
		The values of $x_t$ are standardized to zero mean and unit variance beforehand.
		Triangles denote \DATASETA{}, \DATASETB{} and \DATASETC{}, while circles denote the others.
		According to the CDF, \DATASETA{}, \DATASETB{} and \DATASETC{} have relatively small, medium and large noises.
	}}\label{fig:data-diff-distrib}
\end{figure}
}

We obtain 18 well-maintained business KPIs (where the time span is long enough for training and evaluation) from a large Internet company. All KPIs have an interval of 1 minute between two observations.
\compactsub{
We choose 3 datasets, denoted as \DATASETA{}, \DATASETB{} and \DATASETC{}, according to \cref{fig:data-diff-distrib}, so we can evaluate \DONUT{} for noises at different levels.
}{
We choose 3 datasets, denoted as \DATASETA{}, \DATASETB{} and \DATASETC{}, which have relatively small, medium and large noises among the 18 datasets, so we can evaluate \DONUT{} for noises at different levels.
}
We divide each dataset into training, validation and testing sets, whose ratios are 49\%, 21\%, 30\% respectively.
Figures of datasets \DATASETA{}, \DATASETB{} and \DATASETC{} are shown in \cref{fig:kpi}, while statistics are shown in \cref{tab:details-of-datasets}. The operators of the Internet company labeled all the anomalies in these three datasets. For evaluation purpose, we can consider we have the ground truth of all anomalies in these three datasets.

\begin{table}[htbp]
	\footnotesize
	\centering
	\begin{threeparttable}
		\begin{tabularx}{\columnwidth}{
				p{.3\columnwidth}X
				p{.2\columnwidth}X
				p{.2\columnwidth}X
				p{.2\columnwidth}X
			}
			\toprule
			DataSet & \DATASETA & \DATASETB & \DATASETC   \\
			\midrule
			Total points & 296460 & 317522 & 285120      \\
			Missing points & 1222/0.41\% & 1117/0.35\% & 304/0.11\%           \\
			Anomaly points & 1213/0.41\% & 1883/0.59\% & 4394/1.54\%             \\
			Total windows* & 296341 & 317403 & 285001 \\
			Abnormal windows** & 20460/6.90\% & 20747/6.54\%  & 17288/6.07\%   \\
			\bottomrule
		\end{tabularx}
		\begin{tablenotes}
			\item[*] Each sliding window has a length $W = 120$.
			\item[**] Each abnormal window contains at least one anomaly or missing point.
		\end{tablenotes}
	\end{threeparttable}
	\ifnotcompact
	  \vspace{.5em}
	\fi
	\caption{Statistics of \DATASETA{}, \DATASETB{} and \DATASETC{}.}
	\label{tab:details-of-datasets}
\end{table}


\compactvspace{-1.5em}
\subsection{Performance Metrics}
\label{sec:performance-metrics}

In our evaluation, we totally ignore outputs of all algorithms at missing points (``null'') since they are straightforward to identify.

\begin{figure}
\begin{flushright}
	\includegraphics[width=\columnwidth]{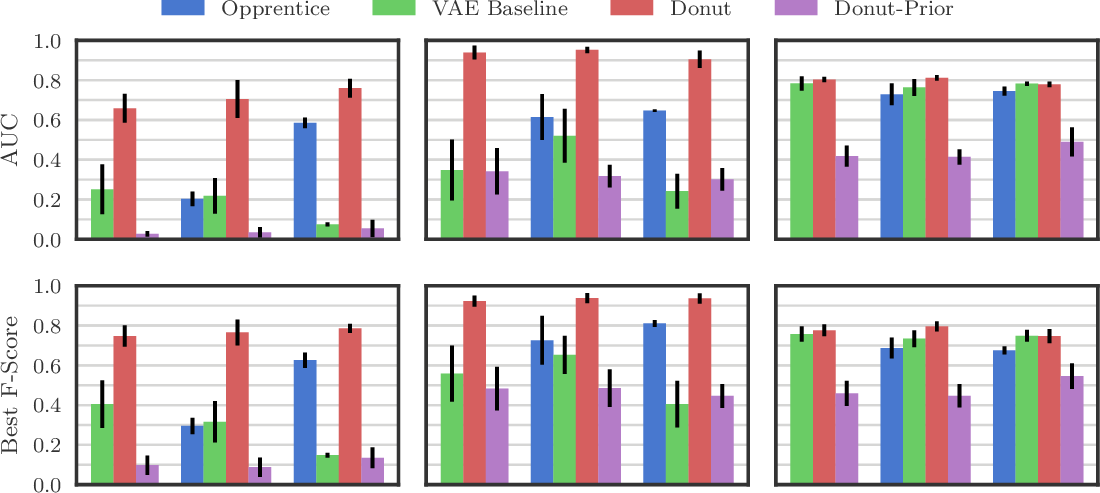}\vspace{.6em}
	\includegraphics[width=\columnwidth]{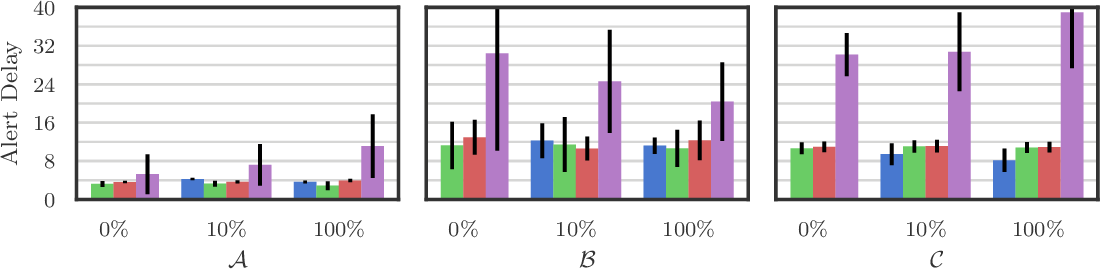}
	\caption{
		AUC, the best F-Score, and the average alert delay corresponding to the best F-score.
		\DATASETA{}, \DATASETB{} and \DATASETC{} are the three datasets.
		``0\%'', ``10\%'' and ``100\%'' are the ratio of the labels preserved in training.  Note there is no result for Opprentice when there are 0\% of anomaly labels.
		The black stick on top of each bar is the deviation of 10 repeated experiments.
	}
	\label{fig:overall-perf}
\end{flushright}
\end{figure}

All the algorithms evaluated in this paper compute one anomaly score for each point. A threshold can be chosen to do the decision: if the score for a point is greater than the threshold, an alert should be triggered.
In this way, anomaly detection is similar to a classification problem, and we may compute the precision and recall corresponding to each threshold.
We may further compute the AUC, which is the average precision over recalls, given all possible thresholds; or the F-score, which is the harmonic mean of precision and recall, given one particular threshold.
We may also enumerate all thresholds, obtaining all F-scores, and use the \textit{best F-score} as the metric.
The best F-score indicates the best possible performance of a model on a particular testing set, given an optimal global threshold.
In practice, the best F-score is mostly consistent with AUC, except for slight differences (see \cref{fig:overall-perf}).
We prefer the best F-score to AUC, since it should be more important to have an excellent F-score at a certain threshold than to have just high but not so excellent F-scores on most thresholds.

\begin{figure}
	\centering
	\ifcompact
	  \includegraphics[width=0.8\columnwidth]{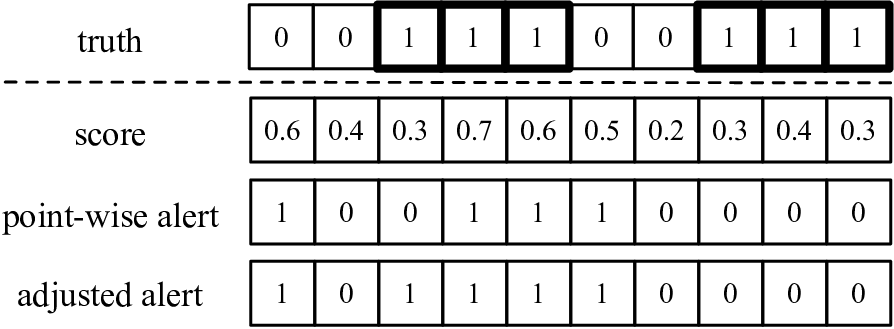}
	\else
	  \includegraphics[width=\columnwidth]{metric-illustration}
	\fi
	\caption{
		Illustration of the strategy for modified metrics. The first row is the truth with 10 contiguous points and two anomaly segments highlighted in the shaded squares. The detector scores are shown in the second row. The third row shows the point-wise detector results with a threshold of 0.5. The forth row shows the detector results after adjustment. We shall get precision 0.6, and recall 0.5. From the third row, the alert delay for the first segment is 1 interval (1 minute).
	}\label{fig:metric-illustration}
\end{figure}

In real applications, the human operators generally do not care about the point-wise metrics.
It is acceptable for an algorithm to trigger an alert for any point in a contiguous anomaly segment, if the delay is not too long.
Some metrics for anomaly detection have been proposed to accommodate this preference, \EG, \cite{evaluation}, but most are not widely accepted, likely because they are too complicated.
We instead use a simple strategy: if any point in an anomaly segment in the ground truth can be detected by a chosen threshold, we say this segment is detected correctly, and all points in this segment are treated as if they can be detected by this threshold.
Meanwhile, the points outside the anomaly segments are treated as usual.
The precision, recall, AUC, F-score and best F-score are then computed accordingly.
This approach is illustrated in \cref{fig:metric-illustration}.

In addition to the accuracy metric, we compute the alert delay for each detected segment, which is also important to  the operators. For a true positive segment, the alert delay is the time difference between the first point and the first detected point in the segment.


\compactvspace{-.5em}
\subsection{Experiment Setup}
\label{sec:experiment-setup}

We set the window size $W$ to be 120, which spans 2 hours in our datasets.
The choice of $W$ is restricted by two factors. On the one hand, too small a $W$ will cause the model to be unable to capture the patterns, since the model is expected to recognize what the normal pattern is with the information only from the window (see \cref{sec:kde-interpretation}). On the other hand, too large a $W$ will increase the risk of over-fitting, since we stick to fully-connected layers without weight sharing, thus the number of model parameters is proportional to $W$.
We set the latent dimension $K$ to be 3 for \DATASETB{} and \DATASETC{}, since the 3-d dimensional space can be easily visualized for analysis and luckily $K=3$ works well empirically for for \DATASETB{} and \DATASETC{}. As for \DATASETA{}, we found 3 is too small, so we empirically increase $K$ to 8.
These empirical choices of $K$ are proven to be quite good on testing set, as will be shown in \cref{fig:z-dim-perf}.
The hidden layers of $q_{\phi}(\vv{z}|\vv{x})$ and $p_{\theta}(\vv{x}|\vv{z})$ are both chosen as two ReLU layers, each with 100 units, which makes the variational and generative network have equal size.
We did not carry out exhaustive search on the structure of hidden networks.

Other hyper-parameters are also chosen empirically.
We use $10^{-4}$ as $\epsilon$ of the std layer.
We use $0.01$ as the injection ratio $\lambda$.
We use 10 as the MCMC iteration count $M$, and use 1024 as the sampling number $L$ of Monte Carlo integration.
We use 256 as the batch size for training, and run for 250 epochs.
We use Adam optimizer~\cite{kingma_adam:_2014}, with an initial learning rate of $10^{-3}$.
We discount the learning rate by 0.75 after every 10 epochs.
We apply L2 regularization to the hidden layers, with a coefficient of $10^{-3}$.
We clip the gradients by norm, with a limit of 10.0.

In order to evaluate \DONUT{} with no labels, we ignore all the labels. For the case of occasional labels, we down-sample the anomaly labels of training and validation set to make it contain 10\% of labeled anomalies. Note that missing points are not down-sampled. We keep throwing away anomaly segments randomly, with a probability that is proportional to the length of each segment, until the desired down-sampling rate is reached. We use this approach instead of randomly throwing away individual anomaly points, because KPIs are time sequences and each anomaly point could leak information about its neighboring points, resulting in over-estimated performance. Such downsampling are done 10 times, which enables us to do 10 independent, repeated experiments. Overall for each dataset, we have three versions: 0\% labels, 10\% labels, and 100\% labels.


\compactvspace{-.7em}
\subsection{Overall Performance}
\label{sec:overall-performance}

We measure the AUC, the best F-Score, and the average alert delay corresponding to the best F-score in \cref{fig:overall-perf} of \DONUT{}, and compared with three selected algorithms.

\textbf{Opprentice~\cite{opprentice}} is an ensemble supervised framework using Random Forest classifier. On datasets similar to ours, Opprentice is reported to consistently and significantly outperform 14 anomaly detectors based on traditional statistical models (\textit{e.g.}, \cite{historical-avg,MA,arma,arima,kalman,holt-winters,TSD,svd,wavelet}), with in total 133 enumerated configurations of hyper-parameters for these detectors. Thus, in our evaluation of \DONUT{}, Opprentice not only serves as a state-of-art competitor algorithm from the non deep learning areas, but also serves as a proxy to compare with the empirical performance ``upper bound" of these traditional anomaly detectors.

\textbf{VAE baseline.} The VAE-based anomaly detection in~\cite{vae-ad} does not deal with time sequences, thus we set up the VAE baseline as follows. First, the VAE baseline has the same network structure as \DONUT{}, as shown in \cref{fig:vae-network-structure}. Second, among all the techniques in \cref{fig:architecture}, only those techniques in the Data Preparation step are used. Third, as suggested by \cite{vae-ad}, we exclude \textbf{all windows} containing either labeled anomalies or missing points from training data.

\textbf{Donut-Prior}. Given that a generative model learns $p(\vv{x})$ by nature, while in VAE $p(\vv{x})$ is defined as $\EEE{p_{\theta}(\vv{z})}{p_{\theta}(\vv{x}|\vv{z})}$, we also evaluate the prior counterpart of reconstruction probability, \IE, $\EEE{p_{\theta}(\vv{z})}{\log p_{\theta}(\vv{x}|\vv{z})}$.  We just need a baseline of the prior, so we compute the prior expectation by plain Monte Carlo integration, without advanced techniques to improve the result.

The best F-score of \DONUT{} is quite satisfactory in totally unsupervised case, ranges from 0.75 to 0.9, better than the supervised Opprentice in all cases.
In fact, when labels are incomplete, the best F-score of the Opprentice drops heavily in \DATASETA{} and \DATASETB{}, only remaining acceptable in \DATASETC{}.
The number of anomalies are much larger in \DATASETC{} than \DATASETA{} and \DATASETB{}, while having 10\% of labels are likely to be just enough for training.
\DONUT{} has an outstanding performance in the unsupervised scenario, and we see that feeding anomaly labels into \DONUT{} would in general make it work even better.
There is, however, an unusual behavior of \DONUT{}, where the best F-score in \DATASETC{}, as well as the AUC in \DATASETB{} and \DATASETC{}, are slightly worse with 100\% labels than 10\%.
This is likely an optimization problem, where the unlabeled anomalies might cause training to be unstable, and accidentally pull the model out of a sub-optimal equilibrium (\cref{sec:sub-optimal-equilibrium}).
Such phenomenon seems to diminish when $K$ increases from 3 (\DATASETB{} and \DATASETC{}) to 8 (\DATASETA{}).
Fortunately, it does not matter too much, so we would suggest to use labels in \DONUT{} whenever possible.

\DONUT{} outperforms the VAE baseline by a large margin in \DATASETA{} and \DATASETB{}, while it does not show such great advantage in \DATASETC{}.
In fact, the relative advantage of \DONUT{} is the largest in \DATASETA{}, medium in \DATASETB{}, and the smallest in \DATASETC{}.
This is caused by the following reasons.
Naturally, VAE models normal $\vv{x}$.
As a result, the reconstruction probability actually expects $\vv{x}$ to be mostly normal (see \cref{sec:kde-interpretation}).
However, since $\vv{x}$ are sliding windows of KPIs and we are required to produce one anomaly score for every point, it is sometimes inevitable to have abnormal points in $\vv{x}$.
This causes the VAE baseline to suffer a lot.
In contrast, the techniques developed in this paper enhances the ability of \DONUT{} to produce reliable outputs even when anomalies present in earlier points in the same window.
Meanwhile,  abnormal points with similar abnormal magnitude would appear relatively ``more abnormal'' when the KPI is smoother. 
Given that \DATASETA{} is the smoothest, \DATASETB{} is medium, and \DATASETC{} is the least smoothest, above observation in the relative advantage is not surprising.

Finally, the best F-score of the \textit{Donut-Prior} is much worse than the reconstruction probability, especially when the dimension of $\vv{z}$ is larger.
However, it is worth mentioning that the posterior expectation in reconstruction probability only works under certain conditions (\cref{sec:find-better-posterior}).
Fortunately, this problem does not matter too much to  \DONUT{} (see \cref{sec:find-better-posterior}). As such, the reconstruction probability can be used without too much concern. 

The average alert delays of \DONUT{}, Opprentice and VAE Baseline are acceptable over all datasets, whereas Donut-Prior is not.  Meanwhile, the best F-score of \DONUT{} is much better than others.  In conclusion, \DONUT{} could achieve the best performance without increasing the alert delay, thus \DONUT{} is practical for operators.  


\compactvspace{-.5em}
\subsection{Effects of \DONUT{} Techniques}
\label{sec:trick-effects}

\begin{figure}
	\centering
	\includegraphics[width=\columnwidth]{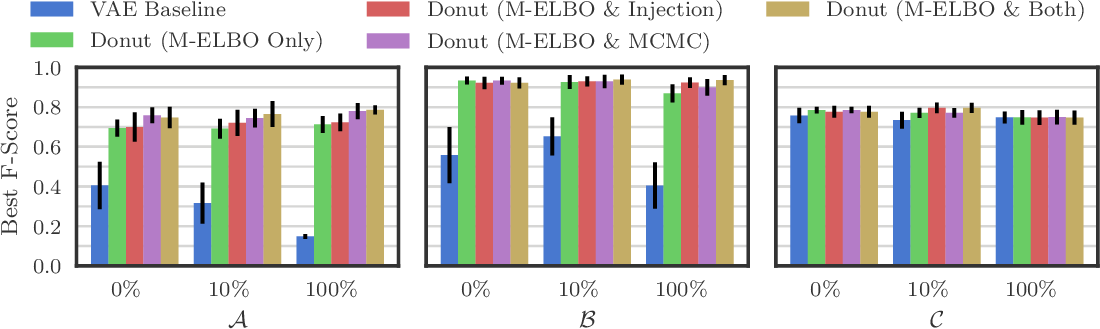}
	\caption{
		Best F-score of (1) VAE baseline, (2) \DONUT{} with M-ELBO, (3) M-ELBO + missing data injection, (4) M-ELBO + MCMC, and (5) M-ELBO + both MCMC and injection.
		The M-ELBO alone contributes most of the improvement.
	}
	\label{fig:tricks-perf}
\end{figure}

We have proposed three techniques in this paper: (1) M-ELBO (\cref{eqn:vae-elbo-modified}), (2) missing data injection, and (3) MCMC imputation.
In \cref{fig:z-dim-perf}, we present the best F-score of \DONUT{} with four possible combinations of these techniques, plus the VAE baseline for comparison.
These techniques are closely related to the KDE interpretation, which will be discussed further in \cref{sec:find-better-posterior}.

\textbf{M-ELBO} alone contributes most of the improvement over the \ifcompact\pagebreak\fi VAE baseline.
It works by training \DONUT{} to get used to possible abnormal points in $\vv{x}$, and to produce desired outputs in such cases.
Although we expected M-ELBO to work, we did not expect it to work such well.
In conclusion, \textbf{it would not be a good practice to train a VAE for anomaly detection using only normal data, although it seems natural for a generative model (\cref{sec:find-better-posterior}).}
To the best of our knowledge, M-ELBO and its importance have never been stated in previous work, thus is a major contribution of ours.

\textbf{Missing data injection} is designed for amplifying the effect of M-ELBO, and can actually be seen as a data augmentation method.
In fact, it would be better if we inject not only missing points, but also synthetically generated anomalies during training.
However, it is difficult to generate anomalies similar enough to the real ones, which should be a large topic and is out of the scope of this paper.
We thus only inject the missing points.
The improvement of best F-score introduced by missing data injection is not very significant, and in the case of 0\% labels on \DATASETB{} and \DATASETC{}, it is slightly worse than M-ELBO only.
This is likely because the injection introduces extra randomness to training, such that it demands larger training epochs, compared to the case of M-ELBO only.
We are not sure how many number of epochs to run when the injection is adopted, in order to get an objective comparison, thus we just use the same epochs in all cases, leaving the result as it is.
We still recommend to use missing data injection, even with a cost of larger training epochs, as it is expected to work with a large chance.

\textbf{MCMC imputation} is also designed to help \DONUT{} deal with abnormal points.
Although \DONUT{} obtains significant improvement of best F-score with MCMC in only some cases, it never harms the performance.
According to~\cite{rezende_stochastic_2014}, this should be an expected result.
We thus recommend to always adopt MCMC in detection.

In conclusion, we recommend to use all the three techniques of \DONUT{}.
The result of such configuration is also presented in \cref{fig:tricks-perf}.


\compactvspace{-1em}
\subsection{Impact of K}
\label{sec:impact-of-k}

\begin{figure}
	\centering
	\includegraphics[width=\columnwidth]{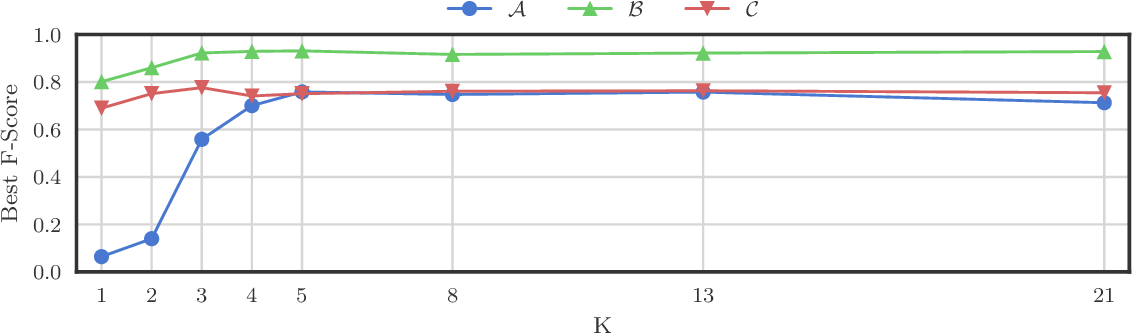}
	\caption{
		The best F-score of unsupervised \DONUT{} with different $K$, averaged over 10 repeated experiments.
	}
	\label{fig:z-dim-perf}
\end{figure}

The number of z dimensions, \IE, $K$, plays an important role.
Too small a $K$ would potentially cause under-fitting, or sub-optimal equilibrium (see \cref{sec:sub-optimal-equilibrium}).
On the other hand, too large a $K$ would probably cause the reconstruction probability unable to find a good posterior (see \cref{sec:kde-interpretation}).
It is difficult to choose a good $K$ in totally unsupervised scenario, thus we leave it as a future work.

In \cref{fig:z-dim-perf},  we present the average best F-score with different $K$ on testing set for unsupervised \DONUT{}.
This does not help us choose the best $K$ (since we cannot use testing test to pick $K$), but can show our empirical choice of $8,3,3$ is quite good. The best F-score reaches maximum at 5 for \DATASETA{}, 4 for \DATASETB{} and 3 for \DATASETC{}.
In other words, the best F-score could be achieved with fairly small $K$.
On the other hand, the best F-score does not drop too heavily for $K$ up to 21.
This gives us a large room to empirically choose $K$.
Finally, we notice that smoother KPIs seem to demand larger $K$.
Such phenomenon is not fully studied in this paper, and we leave it as a future work.
Based on the observations in \cref{fig:z-dim-perf}, for KPIs similar to \DATASETA{}, \DATASETB{} or \DATASETC{}, we suggest an empirical choice of $K$ within the range from 5 to 10.



\compactvspace{-1.7em}
\section{Analysis}
\label{sec:analysis}

\subsection{KDE Interpretation}
\label{sec:kde-interpretation}

\begin{figure}
	\begin{subfigure}[t]{0.31\columnwidth}
		\centering
		\includegraphics[width=\columnwidth]{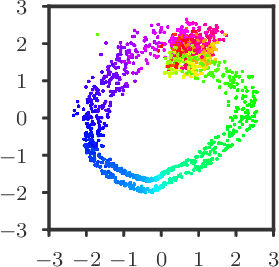}
		\caption{}\label{fig:z2-latent-space-normal}
	\end{subfigure}\hfill
	\begin{subfigure}[t]{0.2675\columnwidth}
		\centering
		\includegraphics[width=\columnwidth]{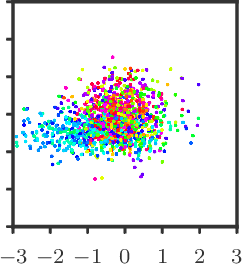}
		\caption{}\label{fig:z2-latent-space-null}
	\end{subfigure}\hfill
	\begin{subfigure}[t]{0.371\columnwidth}
		\centering
		\includegraphics[width=\columnwidth]{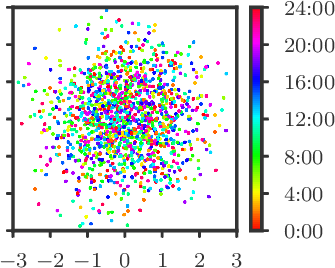}
		\caption{}\label{fig:z2-latent-space-spread-out}
	\end{subfigure}

	\caption{
		The $\vv{z}$ layout of dataset \DATASETB{} with (a) \DONUT{}, (b) untrained VAE, (c) VAE trained using $\EE{\log p_{\theta}(\vv{z})} + \Entropyy{\vv{z}|\vv{x}}$ as loss.
		Figures are plotted by sampling $\vv{z}$ from $q_{\phi}(\vv{z}|\vv{x})$, corresponding to normal $\vv{x}$ randomly chosen from the testing set.
		$K$ is chosen as 2, so the x- and y-axis are the two dimensions of $\vv{z}$ samples.
		We plot $\vv{z}$ samples instead of $\vv{\mu_z}$ of $q_{\phi}(\vv{z}|\vv{x})$, since we want to take into account the effects of $\vv{\sigma_z}$ in the figures.
		The color of $\vv{z}$ a sample denotes its time of the day.
	}
	\label{fig:z2-latent-space}
\end{figure}

\begin{figure*}
	\centering
	\ifcompact
	  \includegraphics[width=0.8\linewidth]{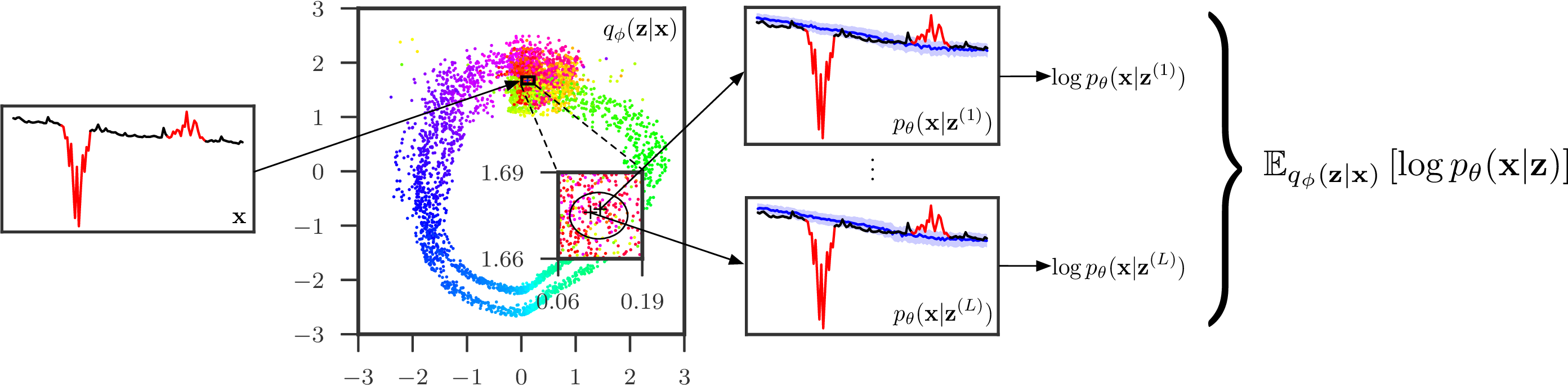}
	\else
	  \includegraphics[width=\linewidth]{kde_interpretation}
	\fi
	\caption{
		Illustration of the KDE interpretation.
		For a given $\vv{x}$ potentially with anomalies, \DONUT{} tries to recognize what normal pattern it follows, encoded as $q_{\phi}(\vv{z}|\vv{x})$.
		The black ellipse in the middle figure denotes the 3-$\vv{\sigma_z}$ region of $q_{\phi}(\vv{z}|\vv{x})$.
		$L$ samples of $\vv{z}$ are then taken from $q_{\phi}(\vv{z}|\vv{x})$, denoted as the crosses in the middle figure.
		Each $\vv{z}$ is associated with a density estimator kernel $\log p_{\theta}(\vv{x}|\vv{z})$.
		The blue curves in the right two figures are $\vv{\mu_x}$ of each kernel, while the surrounding stripes are $\vv{\sigma_x}$.
		Finally, the values of $\log p_{\theta}(\vv{x}|\vv{z})$ are computed from each kernel, and further averaged together as the reconstruction probability.
	}
	\label{fig:kde-interpretation}
\end{figure*}

\begin{figure}
	\centering
	\includegraphics[width=\columnwidth]{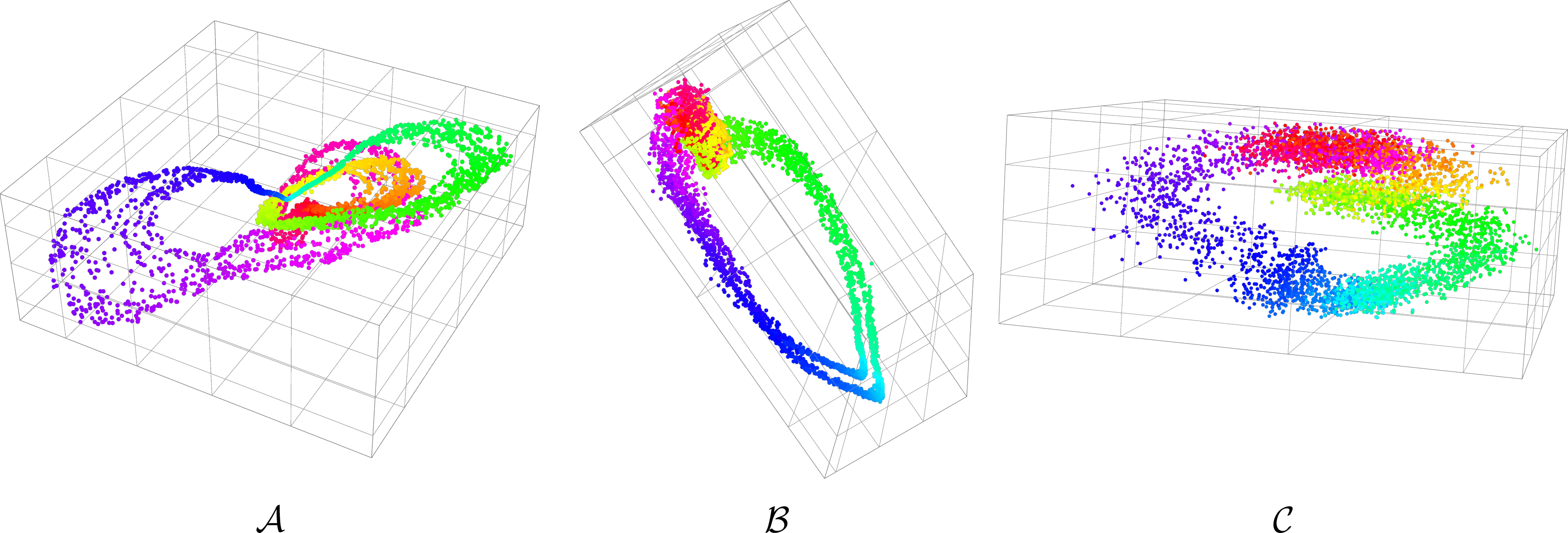}
	\caption{
		3-d latent space of all three datasets.
	}
	\label{fig:z3-latent-space}
\end{figure}

Although the reconstruction probability $\EEE{q_{\phi}(\vv{z}|\vv{x})}{\log p_{\theta}(\vv{x}|\vv{z})}$ has been adopted in \cite{vae-ad,vi-storn}, how it actually works has not yet been made clear.
Some may see it as a variant of $\EEE{q_{\phi}(\vv{z}|\vv{x})}{p_{\theta}(\vv{x}|\vv{z})}$, but $\EEE{q_{\phi}(\vv{z}|\vv{x})}{p_{\theta}(\vv{x}|\vv{z})} = \int p_{\theta}(\vv{x}|\vv{z}) q_{\phi}(\vv{z}|\vv{x}) \dd\vv{z}$, which is definitely not a well-defined probability\footnote{
		In general it should give no useful information by computing the expectation of $\log p_{\theta}(\vv{x}|\vv{z})$ upon the posterior $q_{\phi}(\vv{z}|\vv{x})$, using a \textbf{potentially abnormal} $\vv{x}$.
	}.
Thus neither of \cite{vae-ad,vi-storn} can be explained by the probabilistic framework.
We hereby propose the KDE (kernel density estimation) interpretation for the reconstruction probability, and for the entire \DONUT{} algorithm.

The posterior $q_{\phi}(\vv{z}|\vv{x})$ for normal $\vv{x}$ exhibits time gradient, as \cref{fig:z2-latent-space-normal} shows.
The windows of $\vv{x}$ at contiguous time (\textbf{contiguous $\vv{x}$} for short hereafter) are mapped to nearby $q_{\phi}(\vv{z}|\vv{x})$, mostly with small variance $\vv{\sigma_z}$ (see \cref{fig:kde-interpretation}).
The $q_{\phi}(\vv{z}|\vv{x})$ are thus organized in smooth transition, causing $\vv{z}$ samples to exhibit color gradient in the figure.
We name this structure ``time gradient''.
The KPIs in this paper are smooth in general, so contiguous $\vv{x}$ are highly similar. The root cause of time gradient is the transition of $q_{\phi}(\vv{z}|\vv{x})$ in the shape of $\vv{x}$ (rather than the one in time), because  \DONUT{} consumes only the shape of $\vv{x}$ and no time information.
Time gradient benefits the generalization of \DONUT{} on unseen data: if we have a posterior $q_{\phi}(\vv{z}|\vv{x})$ somewhere between two training posteriors, it would be well-defined, avoiding absurd detection output.

For a partially abnormal $\vv{x}$\footnote{We call a $\vv{x}$ partially abnormal if only a small portion of points within $\vv{x}$ are abnormal, such that we can easily tell what normal pattern $\vv{x}$ should follow.}, the dimension reduction would allow \DONUT{} to recognize its normal pattern $\tilde{\vv{x}}$, and cause $q_{\phi}(\vv{z}|\vv{x})$ to be approximately  $q_{\phi}(\vv{z}|\tilde{\vv{x}})$.
This effect is caused by the following reasons.
\DONUT{} is trained to reconstruct normal points in training samples with best efforts, while the dimension reduction causes \DONUT{} to be only able to capture a small amount of information from $\vv{x}$.
As a result, only the overall shape is encoded in $q_{\phi}(\vv{z}|\vv{x})$.
The abnormal information is likely to be dropped in this procedure.
However, if a $\vv{x}$ is too abnormal, \DONUT{} might fail to recognize any normal $\tilde{\vv{x}}$, such that $q_{\phi}(\vv{z}|\vv{x})$ would become ill-defined.

The fact that $q_{\phi}(\vv{z}|\vv{x})$ for a partially abnormal $\vv{x}$ would be similar to $q_{\phi}(\vv{z}|\tilde{\vv{x}})$ brings special meanings to the reconstruction probability in \DONUT{}.
Since M-ELBO is maximized with regard to normal patterns during training, $\log p_{\theta}(\vv{x}|\vv{z})$ for $\vv{z} \sim q_{\phi}(\vv{z}|\tilde{\vv{x}})$ should produce high scores for $\vv{x}$ similar to $\tilde{\vv{x}}$, and vise versa.
That is to say, each $\log p_{\theta}(\vv{x}|\vv{z})$ can be used as a density estimator, indicating how well $\vv{x}$ follows the normal pattern $\tilde{\vv{x}}$.
The posterior expectation then sums up the scores from all $\log p_{\theta}(\vv{x}|\vv{z})$, with the weight $q_{\phi}(\vv{z}|\vv{x})$ for each $\vv{z}$.
This procedure is very similar to weighted kernel density estimation~\cite{wkde1,wkde2}.
We thus carry out the KDE interpretation: \textbf{the reconstruction probability $\EEE{q_{\phi}(\vv{z}|\vv{x})}{\log p_{\theta}(\vv{x}|\vv{z})}$ in \DONUT{} can be seen as weighted kernel density estimation, with $q_{\phi}(\vv{z}|\vv{x})$ as weights and $\log p_{\theta}(\vv{x}|\vv{z})$ as kernels}
\footnote{The weights $q_{\phi}(\vv{z}|\vv{x})$ are implicitly applied by sampling in Monte Carlo integration.}.

\cref{fig:kde-interpretation} is an illustration of the KDE interpretation.
We also visualize the 3-d latent spaces of all datasets in \cref{fig:z3-latent-space}.
From the KDE interpretation, we suspect the prior expectation would not work well, whatever technique is adopted to improve the result: sampling on the prior should obtain kernels for all patterns of $\vv{x}$, potentially confusing the density estimation for a particular $\vv{x}$.


\compactvspace{-.5em}
\subsection{Find Good Posteriors for Abnormal x}
\label{sec:find-better-posterior}

\DONUT{} can recognize the normal pattern of a partially abnormal $\vv{x}$, and find a good posterior for estimating how well $\vv{x}$ follows the normal pattern. We now analyze how the techniques in \DONUT{} can enhance such ability of finding good posteriors.

\begin{figure}
	\centering
	\ifcompact
	  \includegraphics[width=0.75\columnwidth]{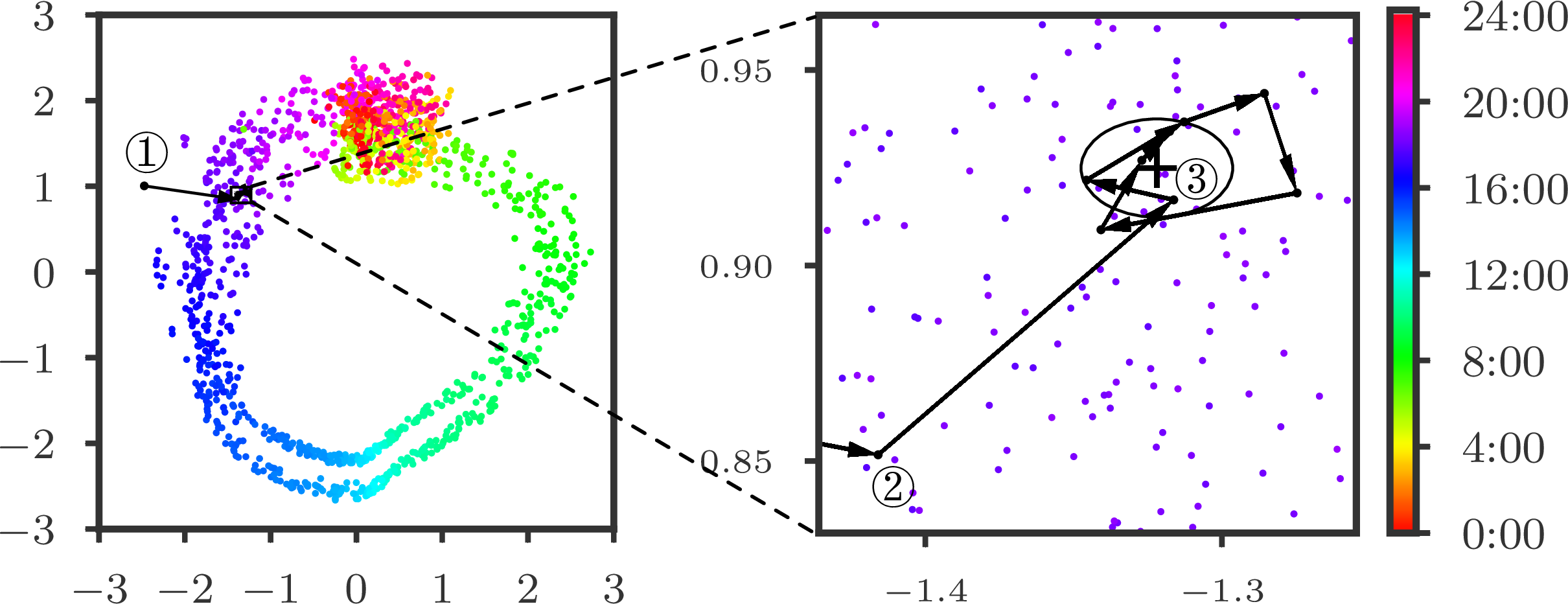}
	\else
	  \includegraphics[width=\columnwidth]{mcmc_z_trace}
	\fi
	\caption{
		MCMC visualization.
		A normal $\vv{x}$ is chosen, whose posterior $q_{\phi}(\vv{z}|\vv{x})$ is plotted at right: the cross denotes $\vv{\mu_x}$ and the ellipse denotes its 3-$\vv{\sigma_x}$ region.
		We randomly set 15\% $\vv{x}$ points as missing, to obtain the abnormal $\vv{x}'$.
		We run MCMC over $\vv{x}'$ with 10 iterations.
		At first, the $\vv{z}$ sample is far from $q_{\phi}(\vv{z}|\vv{x})$.
		After that, $\vv{z}$ samples quickly approach $q_{\phi}(\vv{z}|\vv{x})$, and begin to move around $q_{\phi}(\vv{z}|\vv{x})$ after only 3 iterations.
	}
	\label{fig:mcmc-z-trajectory}
\end{figure}

\DONUT{} is forced to reconstruct normal points within abnormal windows correctly during training, by M-ELBO.
It is thus explicitly trained to find good posteriors.
This is the main reason why M-ELBO plays a vital role in \cref{fig:tricks-perf}.
Missing data injection amplifies the effect of M-ELBO, with synthetically generated missing points.
On the other hand, MCMC imputation does not change the training process.
Instead, it improves the detection, by iteratively approaching better posteriors, as illustrated in \cref{fig:mcmc-z-trajectory}.

Despite these techniques, \DONUT{} may still fail to find a good posterior, if there are too many anomalies in $\vv{x}$.
In our scenario, the KPIs are time sequences, with one point per minute.
For long-lasting anomalies, having the correct detection scores and raise alerts at first few minutes are sufficient in our context\footnote{In practice, ensuing and continuous alerts are typically filtered out anyway.}.
The operators can take action once any score reaches the threshold, and simply ignore the following inaccurate scores.
\textbf{Nevertheless, the KDE interpretation can help us know the limitations of reconstruction probability, in order to use it properly.}


\compactvspace{-.5em}
\subsection{Causes of Time Gradient}
\label{sec:clustering-effect-cause}

In this section we discuss the causes of the time gradient effect.
To simplify the discussion, let us assume training $\vv{x}$ are all normal, thus M-ELBO is now equivalent to the original ELBO. M-ELBO can then be decomposed into three terms as in \cref{eqn:vae-elbo-decomposed} (we leave out some subscripts for shorter notation).
\begin{align*}
	\mathcal{L}(\vv{x})
		&= \EEE{q_{\phi}(\vv{z}|\vv{x})}{
			\log p_{\theta}(\vv{x}|\vv{z}) + \log p_{\theta}(\vv{z}) - \log q_{\phi}(\vv{z}|\vv{x}) } \\
		&= \EE{\log p_{\theta}(\vv{x}|\vv{z})} +
			\EE{\log p_{\theta}(\vv{z})} +
			\Entropyy{\vv{z}|\vv{x}}
			\numberthis\label{eqn:vae-elbo-decomposed}
\end{align*}
The 1st term requires $\vv{z}$ samples from $q_{\phi}(\vv{z}|\vv{x})$ to have a high likelihood of reconstructing $\vv{x}$.
As a result, $q_{\phi}(\vv{z}|\vv{x})$ for $\vv{x}$ with dissimilar shapes are separated.
The 2nd term causes $q_{\phi}(\vv{z}|\vv{x})$ to concentrate on $\mathcal{N}(\vv{0},\vv{I})$.
The 3rd term, the entropy of $q_{\phi}(\vv{z}|\vv{x})$, causes $q_{\phi}(\vv{z}|\vv{x})$ to expand wherever possible.
Recall the 2nd term sets a restricted area for $q_{\phi}(\vv{z}|\vv{x})$ to expand (see \cref{fig:z2-latent-space-spread-out} for the combination effect of the 2nd and 3rd term).
Taking the 1st term into account, this expansion would also stop if $q_{\phi}(\vv{z}|\vv{x})$ for two dissimilar $\vv{x}$ reach each other.
In order for every $q_{\phi}(\vv{z}|\vv{x})$ to have a maximal territory when training converges (\IE, these three terms reach an equilibrium), similar $\vv{x}$ would have to get close to each other, allowing $q_{\phi}(\vv{z}|\vv{x})$ to grow larger with overlapping boundaries.
Since contiguous $\vv{x}$ are similar in seasonal KPIs (and vise versa), the time gradient would be a natural consequence, if such equilibrium could be achieved.

\begin{figure}
	\centering
	\ifcompact
      \includegraphics[height=0.6in]{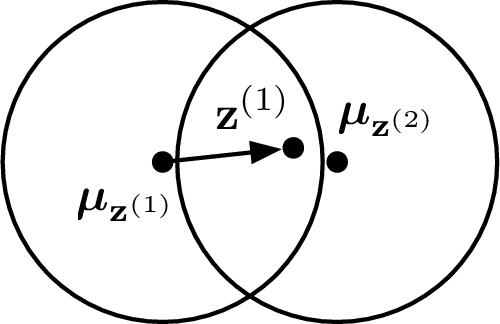}
    \else
      \includegraphics[height=1.0in]{sgvb-dynamic}
    \fi
    \caption{
    	Suppose $\vv{\mu}_{\vv{z}^{(1)}}$ and $\vv{\mu}_{\vv{z}^{(2)}}$ are the mean of $q_{\phi}(\vv{z}|\vv{x})$ corresponding to training data $\vv{x}^{(1)}$ and $\vv{x}^{(2)}$, with the surrounding circles represent $\vv{\sigma}_{\vv{z}^{(1)}}$ and $\vv{\sigma}_{\vv{z}^{(2)}}$.
    	When these two distributions accidentally ``overlaps'' during training, the sample $\vv{z}^{(1)}$ from $q_{\phi}(\vv{z}|\vv{x}^{(1)})$ may get too close to $\vv{\mu}_{\vv{z}^{(2)}}$, such that the reconstructed distribution will be close to $p_{\theta}(\vv{x}|\vv{z}^{(2)})$ with some $\vv{z}^{(2)}$ for $\vv{x}^{(2)}$.
    	If $\vv{x}^{(1)}$ and $\vv{x}^{(2)}$ are dissimilar, $\log p_{\theta}(\vv{x}^{(1)}|\vv{z}^{(2)})$ in the loss will then effectively push $\vv{\mu}_{\vv{z}^{(1)}}$ away from $\vv{\mu}_{\vv{z}^{(2)}}$.
    }\label{fig:sgvb-dynamic}
\end{figure}

\begin{figure}
	\begin{subfigure}[t]{\columnwidth}
		\centering
		\includegraphics[width=\columnwidth]{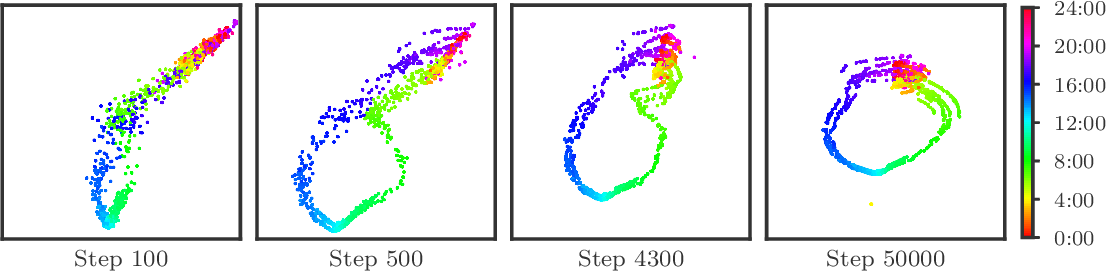}
		\caption{}\label{fig:training-dynamics-good}
	\end{subfigure}
	\begin{subfigure}[t]{\columnwidth}
		\centering
		\includegraphics[width=\columnwidth]{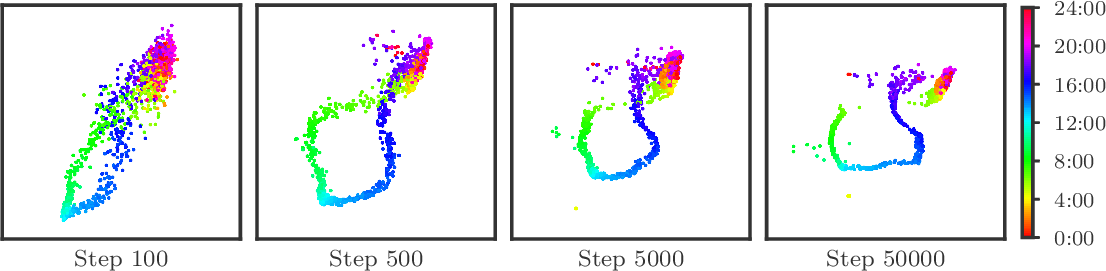}
		\caption{}\label{fig:training-dynamics-bad}
	\end{subfigure}
	\caption{
		Evolution of the $\vv{z}$ space of dataset \DATASETB{} during training.
		We sample normal $\vv{x}$ from validation set, and plot $\vv{z}$ samples accordingly.
		(a) converges to a good equilibrium, with a final F-score 0.871, while (b) converges to a sub-optimal one, with a final F-score 0.826.
		We plot step 4300 in (a), because it is a very important turning point, where the green points just begin to get away from the purple points.
	}
	\label{fig:training-dynamics}
\end{figure}

Next we discuss how the equilibrium could be achieved.
The SGVB algorithm keeps pushing $q_{\phi}(\vv{z}|\vv{x})$ for dissimilar $\vv{x}$ away during training, as illustrated in \cref{fig:sgvb-dynamic}.
The more dissimilar two $q_{\phi}(\vv{z}|\vv{x})$ are, the further they are pushed away.
Since we initialize the variational network randomly, $q_{\phi}(\vv{z}|\vv{x})$ are mixed everywhere when training just begins, as \cref{fig:z2-latent-space-null} shows.
At this time, every $q_{\phi}(\vv{z}|\vv{x})$ are pushed away by all other $q_{\phi}(\vv{z}|\vv{x})$.
Since $\vv{x}$ are sliding windows of KPIs, any pair of $\vv{x}$ far away in time will be generally more dissimilar, thus get pushed away further from each other.
This gives $q_{\phi}(\vv{z}|\vv{x})$ an initial layout.
As training goes on, the time gradient is fine-tuned and gradually established, as \cref{fig:training-dynamics-good} shows.
The training dynamics also suggest that the learning rate annealing technique is very important, since it can gradually stabilize the layout.

Surprisingly, we cannot find any term in M-ELBO that directly pulls $q_{\phi}(\vv{z}|\vv{x})$ for similar $\vv{x}$ together.
The time gradient is likely to be caused mainly by expansion ($\Entropyy{\vv{z}|\vv{x}}$), squeezing ($\EE{\log p_{\theta}(\vv{z})}$), pushing ($\EE{\log p_{\theta}(\vv{x}|\vv{z})}$), and the training dynamics (random initialization and SGVB).
This could sometimes cause trouble, and result in sub-optimal layouts, as we shall see in \cref{sec:sub-optimal-equilibrium}.


\compactvspace{-.5em}
\subsection{Sub-Optimal Equilibrium}
\label{sec:sub-optimal-equilibrium}

\compactdel{
\begin{figure}
	\centering
	\includegraphics[width=\columnwidth]{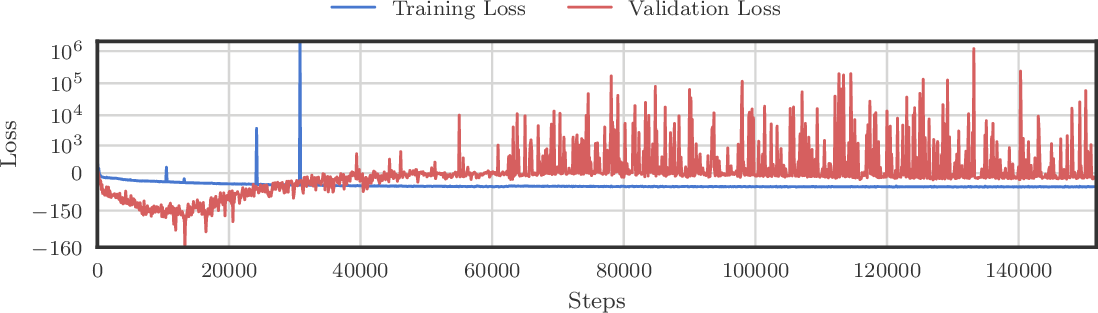}
	\caption{
		Training and validation loss of \cref{fig:training-dynamics-bad}.
	}
	\label{fig:training-dynamics-bad-loss}
\end{figure}
}

$q_{\phi}(\vv{z}|\vv{x})$ may sometimes converge to a sub-optimal equilibrium.
\cref{fig:training-dynamics-bad} demonstrates such a problem, where the purple points accidentally get through the green points after the first 100 steps.
The purple points push the green points away towards both sides, causing the green points to be totally cut off at around 5000 steps.
As training goes on, the green points will be pushed even further, such that the model is locked to this sub-optimal equilibrium and never escapes.
\compactdel{
\cref{fig:training-dynamics-bad-loss} plots the training and validation loss of \cref{fig:training-dynamics-bad}.
Clearly, the model begins to over-fit soon after the green points are separated into halves.
}
Such bad layout of $\vv{z}$ breaks the time gradient, where a testing $\vv{x}$ following green patterns might accidentally be mapped to somewhere between the green two halves and get recognized as purple.
This would certainly downgrade the detection performance, according to the KDE interpretation.

When there are unlabeled anomalies, the training would become unstable so that the model might be accidentally brought out of a sub-optimal equilibrium and achieve a better equilibrium afterwards.
With the help of early-stopping during training, the best encountered equilibrium is chosen eventually.
This explains why sometimes having complete labels would not benefit the performance.
This effect is likely to be less obvious with larger $K$, since having more dimensions gives $q_{\phi}(\vv{z}|\vv{x})$ extra freedom to grow, reducing the chance of  bad layouts.
When sub-optimal equilibrium is not a vital problem, the convergence of training then becomes more important, while having more labels definitely helps stabilize the training.
In conclusion, using anomaly labels in \DONUT{} is likely to benefit the performance, as long as $K$ is adequately large.



\ifnotcompact
\section{Discussion}
\label{sec:discussion}

\subsection{Broader Implications}
\label{sec:implication}

\compactdel{
\textbf{Reconstruction.} The dimension reduction in \DONUT{} throws away information of abnormal points.
This implies that $\vv{z}$ for abnormal $\vv{x}$ might be indistinguishable from normal ones, thus cannot be used directly to detect anomalies.
As a result, the reconstruction is thus an essential step in \DONUT{}.
We believe this conclusion is also true in other algorithms involving dimension reduction. For example, the performance of  PCA-based anomaly detection~\cite{pca_ad1,pca_ad2,sensitivity_pca} 
is sensitive to the number of principle components. We suspect that using the reconstructed samples (as done in \DONUT{}) from these components could remove this sensitivity.
}

\compactdel{
\textbf{KDE interpretation} is the heart of \DONUT{}.  We suspect this interpretation can also benefit the design of other deep generative models in anomaly detection.  Meanwhile, the techniques of this paper (\IE, M-ELBO, missing data injection and MCMC) are designed to enhance the ability of finding good posteriors according to abnormal $\vv{x}$, needed by the density estimation step.  These techniques are also readily applicable to other deep generative models.
}

\compactdel{
\textbf{The time gradient effect} may take place in more general types of seasonal or periodical sequences, rather than just the seasonal KPIs.
Similar $\vv{x}$ are mapped to neighborhood $q_{\phi}(\vv{z}|\vv{x})$, such that \DONUT{} or its variants might potentially be useful in tasks dealing with similarities of sequences, in addition to anomaly detection, \EG, retrieving similar curves from a large database.
}


\subsection{Future Work}
\label{sec:limitation}
%

As mentioned in \cref{sec:impact-of-k}, we leave the task of choosing $K$ as future work.
The topology of $\vv{z}$ might be a key for solving this problem, as we have seen how sub-optimal equilibriums might affect the performance (see \cref{sec:sub-optimal-equilibrium}).
Also, we suspect the sub-optimal equilibriums might be less likely to take place with large $K$, although we may still need more experiments and analysis to prove this.

We did not discuss how to choose the right threshold for detection.
This is also a quite difficult problem, especially in the unsupervised scenario.
Some work (\EG, ~\cite{egads}) have been proposed \WRT{} choosing a proper threshold, which might be applicable to \DONUT{}.

More complicated architectures may be adopted to extend \DONUT{}.
For example, the sequence-to-sequence RNN architecture~\cite{RNN} may be used to replace the fully connected layers in \DONUT{}, so as to handle larger windows, and to deal better with the correlations across points.



\else
  \ifdevelop
\section{Discussion}
\label{sec:discussion}

\subsection{Broader Implications}
\label{sec:implication}

\compactdel{
\textbf{Reconstruction.} The dimension reduction in \DONUT{} throws away information of abnormal points.
This implies that $\vv{z}$ for abnormal $\vv{x}$ might be indistinguishable from normal ones, thus cannot be used directly to detect anomalies.
As a result, the reconstruction is thus an essential step in \DONUT{}.
We believe this conclusion is also true in other algorithms involving dimension reduction. For example, the performance of  PCA-based anomaly detection~\cite{pca_ad1,pca_ad2,sensitivity_pca} 
is sensitive to the number of principle components. We suspect that using the reconstructed samples (as done in \DONUT{}) from these components could remove this sensitivity.
}

\compactdel{
\textbf{KDE interpretation} is the heart of \DONUT{}.  We suspect this interpretation can also benefit the design of other deep generative models in anomaly detection.  Meanwhile, the techniques of this paper (\IE, M-ELBO, missing data injection and MCMC) are designed to enhance the ability of finding good posteriors according to abnormal $\vv{x}$, needed by the density estimation step.  These techniques are also readily applicable to other deep generative models.
}

\compactdel{
\textbf{The time gradient effect} may take place in more general types of seasonal or periodical sequences, rather than just the seasonal KPIs.
Similar $\vv{x}$ are mapped to neighborhood $q_{\phi}(\vv{z}|\vv{x})$, such that \DONUT{} or its variants might potentially be useful in tasks dealing with similarities of sequences, in addition to anomaly detection, \EG, retrieving similar curves from a large database.
}


\subsection{Future Work}
\label{sec:limitation}
%

As mentioned in \cref{sec:impact-of-k}, we leave the task of choosing $K$ as future work.
The topology of $\vv{z}$ might be a key for solving this problem, as we have seen how sub-optimal equilibriums might affect the performance (see \cref{sec:sub-optimal-equilibrium}).
Also, we suspect the sub-optimal equilibriums might be less likely to take place with large $K$, although we may still need more experiments and analysis to prove this.

We did not discuss how to choose the right threshold for detection.
This is also a quite difficult problem, especially in the unsupervised scenario.
Some work (\EG, ~\cite{egads}) have been proposed \WRT{} choosing a proper threshold, which might be applicable to \DONUT{}.

More complicated architectures may be adopted to extend \DONUT{}.
For example, the sequence-to-sequence RNN architecture~\cite{RNN} may be used to replace the fully connected layers in \DONUT{}, so as to handle larger windows, and to deal better with the correlations across points.



  \fi
\fi

\compactvspace{-.3em}
\section{Conclusion}
\label{sec:conclusion}

In this paper, we proposed an unsupervised anomaly detection algorithm  \DONUT{} based on VAE for seasonal KPIs with local variations. The new techniques enabled \DONUT{} to greatly outperform state-of-art supervised and VAE-based anomaly detection algorithms.  The best F-scores of \DONUT{} range from 0.75 to 0.90 for the studied KPIs.

\DONUT{}'s excellent performance are explained by our theoretical analysis with KDE interpretation and the new discovery of the time gradient effect. Our experimental and theoretical analyses imply broader impacts: anomaly detection based on dimension reduction needs to use reconstruction; anomaly detection with generative models needs to train with both normal and abnormal data.


\compactvspace{-.3em}
\section{Acknowledgements}

The work was supported by National Natural Science Foundation of China (NSFC) under grant  No. 61472214 and No. 61472210, and Alibaba Innovative Research (AIR).
We also thank Prof. Jun Zhu and his PhD. student Jiaxin Shi for helpful and constructive discussions.




\newpage
\bibliographystyle{ACM-Reference-Format}
\bibliography{references}


\begin{thebibliography}{41}


\ifx \showCODEN    \undefined \def \showCODEN     #1{\unskip}     \fi
\ifx \showDOI      \undefined \def \showDOI       #1{#1}\fi
\ifx \showISBNx    \undefined \def \showISBNx     #1{\unskip}     \fi
\ifx \showISBNxiii \undefined \def \showISBNxiii  #1{\unskip}     \fi
\ifx \showISSN     \undefined \def \showISSN      #1{\unskip}     \fi
\ifx \showLCCN     \undefined \def \showLCCN      #1{\unskip}     \fi
\ifx \shownote     \undefined \def \shownote      #1{#1}          \fi
\ifx \showarticletitle \undefined \def \showarticletitle #1{#1}   \fi
\ifx \showURL      \undefined \def \showURL       {\relax}        \fi
\providecommand\bibfield[2]{#2}
\providecommand\bibinfo[2]{#2}
\providecommand\natexlab[1]{#1}
\providecommand\showeprint[2][]{arXiv:#2}

\bibitem[\protect\citeauthoryear{Amer, Goldstein, and Abdennadher}{Amer
  et~al\mbox{.}}{2013}]%
        {one-class-svm1}
\bibfield{author}{\bibinfo{person}{Mennatallah Amer}, \bibinfo{person}{Markus
  Goldstein}, {and} \bibinfo{person}{Slim Abdennadher}.}
  \bibinfo{year}{2013}\natexlab{}.
\newblock \showarticletitle{Enhancing one-class support vector machines for
  unsupervised anomaly detection}. In \bibinfo{booktitle}{\emph{Proceedings of
  the ACM SIGKDD Workshop on Outlier Detection and Description}}. ACM,
  \bibinfo{pages}{8--15}.
\newblock


\bibitem[\protect\citeauthoryear{An and Cho}{An and Cho}{2015}]%
        {vae-ad}
\bibfield{author}{\bibinfo{person}{Jinwon An} {and} \bibinfo{person}{Sungzoon
  Cho}.} \bibinfo{year}{2015}\natexlab{}.
\newblock \bibinfo{booktitle}{\emph{Variational Autoencoder based Anomaly
  Detection using Reconstruction Probability}}.
\newblock \bibinfo{type}{{T}echnical {R}eport}. \bibinfo{institution}{SNU Data
  Mining Center}. \bibinfo{pages}{1--18} pages.
\newblock


\bibitem[\protect\citeauthoryear{Beal}{Beal}{2003}]%
        {variational}
\bibfield{author}{\bibinfo{person}{Matthew~James Beal}.}
  \bibinfo{year}{2003}\natexlab{}.
\newblock \bibinfo{booktitle}{\emph{Variational algorithms for approximate
  Bayesian inference}}.
\newblock \bibinfo{publisher}{University of London London}.
\newblock


\bibitem[\protect\citeauthoryear{Bishop}{Bishop}{2006}]%
        {prml}
\bibfield{author}{\bibinfo{person}{Christopher~M Bishop}.}
  \bibinfo{year}{2006}\natexlab{}.
\newblock \bibinfo{booktitle}{\emph{Pattern recognition and machine learning}}.
\newblock \bibinfo{publisher}{springer}.
\newblock


\bibitem[\protect\citeauthoryear{Chandola, Banerjee, and Kumar}{Chandola
  et~al\mbox{.}}{2009}]%
        {ad-survey}
\bibfield{author}{\bibinfo{person}{Varun Chandola}, \bibinfo{person}{Arindam
  Banerjee}, {and} \bibinfo{person}{Vipin Kumar}.}
  \bibinfo{year}{2009}\natexlab{}.
\newblock \showarticletitle{Anomaly detection: A survey}.
\newblock \bibinfo{journal}{\emph{ACM computing surveys (CSUR)}}
  \bibinfo{volume}{41}, \bibinfo{number}{3} (\bibinfo{year}{2009}),
  \bibinfo{pages}{15}.
\newblock


\bibitem[\protect\citeauthoryear{Chen, Mahajan, Sridharan, and Zhang}{Chen
  et~al\mbox{.}}{2013}]%
        {TSD}
\bibfield{author}{\bibinfo{person}{Yingying Chen}, \bibinfo{person}{Ratul
  Mahajan}, \bibinfo{person}{Baskar Sridharan}, {and} \bibinfo{person}{Zhi-Li
  Zhang}.} \bibinfo{year}{2013}\natexlab{}.
\newblock \showarticletitle{A Provider-side View of Web Search Response Time}.
  In \bibinfo{booktitle}{\emph{Proceedings of the ACM SIGCOMM 2013 Conference
  on SIGCOMM}} \emph{(\bibinfo{series}{SIGCOMM '13})}.
  \bibinfo{publisher}{ACM}, \bibinfo{address}{New York, NY, USA},
  \bibinfo{pages}{243--254}.
\newblock
\showISBNx{978-1-4503-2056-6}
\urldef\tempurl%
\url{https://doi.org/10.1145/2486001.2486035}
\showDOI{\tempurl}


\bibitem[\protect\citeauthoryear{Erfani, Rajasegarar, Karunasekera, and
  Leckie}{Erfani et~al\mbox{.}}{2016}]%
        {deep-learning-svm}
\bibfield{author}{\bibinfo{person}{Sarah~M Erfani}, \bibinfo{person}{Sutharshan
  Rajasegarar}, \bibinfo{person}{Shanika Karunasekera}, {and}
  \bibinfo{person}{Christopher Leckie}.} \bibinfo{year}{2016}\natexlab{}.
\newblock \showarticletitle{High-dimensional and large-scale anomaly detection
  using a linear one-class SVM with deep learning}.
\newblock \bibinfo{journal}{\emph{Pattern Recognition}}  \bibinfo{volume}{58}
  (\bibinfo{year}{2016}), \bibinfo{pages}{121--134}.
\newblock


\bibitem[\protect\citeauthoryear{Fontugne, Borgnat, Abry, and Fukuda}{Fontugne
  et~al\mbox{.}}{2010}]%
        {majority_vote}
\bibfield{author}{\bibinfo{person}{Romain Fontugne}, \bibinfo{person}{Pierre
  Borgnat}, \bibinfo{person}{Patrice Abry}, {and} \bibinfo{person}{Kensuke
  Fukuda}.} \bibinfo{year}{2010}\natexlab{}.
\newblock \showarticletitle{MAWILab: Combining Diverse Anomaly Detectors for
  Automated Anomaly Labeling and Performance Benchmarking}. In
  \bibinfo{booktitle}{\emph{Proceedings of the 6th International COnference}}
  \emph{(\bibinfo{series}{Co-NEXT '10})}. \bibinfo{publisher}{ACM}, Article
  \bibinfo{articleno}{8}, \bibinfo{numpages}{12}~pages.
\newblock
\showISBNx{978-1-4503-0448-1}
\urldef\tempurl%
\url{https://doi.org/10.1145/1921168.1921179}
\showDOI{\tempurl}


\bibitem[\protect\citeauthoryear{Fu, Hu, and Tan}{Fu et~al\mbox{.}}{2005}]%
        {cluster}
\bibfield{author}{\bibinfo{person}{Zhouyu Fu}, \bibinfo{person}{Weiming Hu},
  {and} \bibinfo{person}{Tieniu Tan}.} \bibinfo{year}{2005}\natexlab{}.
\newblock \showarticletitle{Similarity based vehicle trajectory clustering and
  anomaly detection}. In \bibinfo{booktitle}{\emph{Image Processing, 2005. ICIP
  2005. IEEE International Conference on}}, Vol.~\bibinfo{volume}{2}. IEEE,
  \bibinfo{pages}{II--602}.
\newblock


\bibitem[\protect\citeauthoryear{Geweke}{Geweke}{1989}]%
        {geweke1989bayesian}
\bibfield{author}{\bibinfo{person}{John Geweke}.}
  \bibinfo{year}{1989}\natexlab{}.
\newblock \showarticletitle{Bayesian inference in econometric models using
  Monte Carlo integration}.
\newblock \bibinfo{journal}{\emph{Econometrica: Journal of the Econometric
  Society}} (\bibinfo{year}{1989}), \bibinfo{pages}{1317--1339}.
\newblock


\bibitem[\protect\citeauthoryear{Gisbert}{Gisbert}{2003}]%
        {wkde1}
\bibfield{author}{\bibinfo{person}{Francisco J~Goerlich Gisbert}.}
  \bibinfo{year}{2003}\natexlab{}.
\newblock \showarticletitle{Weighted samples, kernel density estimators and
  convergence}.
\newblock \bibinfo{journal}{\emph{Empirical Economics}} \bibinfo{volume}{28},
  \bibinfo{number}{2} (\bibinfo{year}{2003}), \bibinfo{pages}{335--351}.
\newblock


\bibitem[\protect\citeauthoryear{Goodfellow, Bengio, and Courville}{Goodfellow
  et~al\mbox{.}}{2016}]%
        {deep_learning_book}
\bibfield{author}{\bibinfo{person}{Ian Goodfellow}, \bibinfo{person}{Yoshua
  Bengio}, {and} \bibinfo{person}{Aaron Courville}.}
  \bibinfo{year}{2016}\natexlab{}.
\newblock \bibinfo{booktitle}{\emph{Deep Learning}}.
\newblock \bibinfo{publisher}{MIT Press}.
\newblock


\bibitem[\protect\citeauthoryear{Goodfellow, Pouget-Abadie, Mirza, Xu,
  Warde-Farley, Ozair, Courville, and Bengio}{Goodfellow et~al\mbox{.}}{2014}]%
        {gan}
\bibfield{author}{\bibinfo{person}{Ian Goodfellow}, \bibinfo{person}{Jean
  Pouget-Abadie}, \bibinfo{person}{Mehdi Mirza}, \bibinfo{person}{Bing Xu},
  \bibinfo{person}{David Warde-Farley}, \bibinfo{person}{Sherjil Ozair},
  \bibinfo{person}{Aaron Courville}, {and} \bibinfo{person}{Yoshua Bengio}.}
  \bibinfo{year}{2014}\natexlab{}.
\newblock \showarticletitle{Generative adversarial nets}. In
  \bibinfo{booktitle}{\emph{Advances in neural information processing
  systems}}. \bibinfo{pages}{2672--2680}.
\newblock


\bibitem[\protect\citeauthoryear{H{\"a}rdle, Werwatz, M{\"u}ller, and
  Sperlich}{H{\"a}rdle et~al\mbox{.}}{2004}]%
        {wkde2}
\bibfield{author}{\bibinfo{person}{Wolfgang H{\"a}rdle}, \bibinfo{person}{Axel
  Werwatz}, \bibinfo{person}{Marlene M{\"u}ller}, {and} \bibinfo{person}{Stefan
  Sperlich}.} \bibinfo{year}{2004}\natexlab{}.
\newblock \showarticletitle{Nonparametric density estimation}.
\newblock \bibinfo{journal}{\emph{Nonparametric and Semiparametric Models}}
  (\bibinfo{year}{2004}), \bibinfo{pages}{39--83}.
\newblock


\bibitem[\protect\citeauthoryear{Kingma and Ba}{Kingma and Ba}{2014}]%
        {kingma_adam:_2014}
\bibfield{author}{\bibinfo{person}{Diederik Kingma} {and}
  \bibinfo{person}{Jimmy Ba}.} \bibinfo{year}{2014}\natexlab{}.
\newblock \showarticletitle{Adam: {A} method for stochastic optimization}.
\newblock \bibinfo{journal}{\emph{arXiv preprint arXiv:1412.6980}}
  (\bibinfo{year}{2014}).
\newblock


\bibitem[\protect\citeauthoryear{Kingma and Welling}{Kingma and
  Welling}{2014}]%
        {kingma_auto-encoding_2014}
\bibfield{author}{\bibinfo{person}{Diederik~P Kingma} {and}
  \bibinfo{person}{Max Welling}.} \bibinfo{year}{2014}\natexlab{}.
\newblock \showarticletitle{Auto-{Encoding} {Variational} {Bayes}}. In
  \bibinfo{booktitle}{\emph{Proceedings of the {International} {Conference} on
  {Learning} {Representations}}}.
\newblock


\bibitem[\protect\citeauthoryear{Knorn and Leith}{Knorn and Leith}{2008}]%
        {kalman}
\bibfield{author}{\bibinfo{person}{Florian Knorn} {and}
  \bibinfo{person}{Douglas~J Leith}.} \bibinfo{year}{2008}\natexlab{}.
\newblock \showarticletitle{Adaptive kalman filtering for anomaly detection in
  software appliances}. In \bibinfo{booktitle}{\emph{INFOCOM Workshops 2008,
  IEEE}}. IEEE, \bibinfo{pages}{1--6}.
\newblock


\bibitem[\protect\citeauthoryear{Krishnamurthy, Sen, Zhang, and
  Chen}{Krishnamurthy et~al\mbox{.}}{2003}]%
        {MA}
\bibfield{author}{\bibinfo{person}{Balachander Krishnamurthy},
  \bibinfo{person}{Subhabrata Sen}, \bibinfo{person}{Yin Zhang}, {and}
  \bibinfo{person}{Yan Chen}.} \bibinfo{year}{2003}\natexlab{}.
\newblock \showarticletitle{Sketch-based change detection: methods, evaluation,
  and applications}. In \bibinfo{booktitle}{\emph{Proceedings of the 3rd ACM
  SIGCOMM conference on Internet measurement}}. ACM, \bibinfo{pages}{234--247}.
\newblock


\bibitem[\protect\citeauthoryear{Lakhina, Crovella, and Diot}{Lakhina
  et~al\mbox{.}}{2004}]%
        {pca_ad1}
\bibfield{author}{\bibinfo{person}{Anukool Lakhina}, \bibinfo{person}{Mark
  Crovella}, {and} \bibinfo{person}{Christophe Diot}.}
  \bibinfo{year}{2004}\natexlab{}.
\newblock \showarticletitle{Diagnosing Network-wide Traffic Anomalies}. In
  \bibinfo{booktitle}{\emph{Proceedings of the 2004 Conference on Applications,
  Technologies, Architectures, and Protocols for Computer Communications}}
  \emph{(\bibinfo{series}{SIGCOMM '04})}. \bibinfo{publisher}{ACM},
  \bibinfo{address}{New York, NY, USA}, \bibinfo{pages}{219--230}.
\newblock
\showISBNx{1-58113-862-8}
\urldef\tempurl%
\url{https://doi.org/10.1145/1015467.1015492}
\showDOI{\tempurl}


\bibitem[\protect\citeauthoryear{Lakhina, Crovella, and Diot}{Lakhina
  et~al\mbox{.}}{2005}]%
        {pca_ad2}
\bibfield{author}{\bibinfo{person}{Anukool Lakhina}, \bibinfo{person}{Mark
  Crovella}, {and} \bibinfo{person}{Christophe Diot}.}
  \bibinfo{year}{2005}\natexlab{}.
\newblock \showarticletitle{Mining Anomalies Using Traffic Feature
  Distributions}. In \bibinfo{booktitle}{\emph{Proceedings of the 2005
  Conference on Applications, Technologies, Architectures, and Protocols for
  Computer Communications}} \emph{(\bibinfo{series}{SIGCOMM '05})}.
  \bibinfo{publisher}{ACM}, \bibinfo{address}{New York, NY, USA},
  \bibinfo{pages}{217--228}.
\newblock
\showISBNx{1-59593-009-4}
\urldef\tempurl%
\url{https://doi.org/10.1145/1080091.1080118}
\showDOI{\tempurl}


\bibitem[\protect\citeauthoryear{Laptev, Amizadeh, and Flint}{Laptev
  et~al\mbox{.}}{2015}]%
        {egads}
\bibfield{author}{\bibinfo{person}{Nikolay Laptev}, \bibinfo{person}{Saeed
  Amizadeh}, {and} \bibinfo{person}{Ian Flint}.}
  \bibinfo{year}{2015}\natexlab{}.
\newblock \showarticletitle{Generic and scalable framework for automated
  time-series anomaly detection}. In \bibinfo{booktitle}{\emph{Proceedings of
  the 21th ACM SIGKDD International Conference on Knowledge Discovery and Data
  Mining}}. ACM, \bibinfo{pages}{1939--1947}.
\newblock


\bibitem[\protect\citeauthoryear{Lavin and Ahmad}{Lavin and Ahmad}{2015}]%
        {evaluation}
\bibfield{author}{\bibinfo{person}{Alexander Lavin} {and}
  \bibinfo{person}{Subutai Ahmad}.} \bibinfo{year}{2015}\natexlab{}.
\newblock \showarticletitle{Evaluating Real-Time Anomaly Detection
  Algorithms--The Numenta Anomaly Benchmark}. In
  \bibinfo{booktitle}{\emph{Machine Learning and Applications (ICMLA), 2015
  IEEE 14th International Conference on}}. IEEE, \bibinfo{pages}{38--44}.
\newblock


\bibitem[\protect\citeauthoryear{Laxhammar, Falkman, and Sviestins}{Laxhammar
  et~al\mbox{.}}{2009}]%
        {GMM}
\bibfield{author}{\bibinfo{person}{Rikard Laxhammar}, \bibinfo{person}{Goran
  Falkman}, {and} \bibinfo{person}{Egils Sviestins}.}
  \bibinfo{year}{2009}\natexlab{}.
\newblock \showarticletitle{Anomaly detection in sea traffic-a comparison of
  the gaussian mixture model and the kernel density estimator}. In
  \bibinfo{booktitle}{\emph{Information Fusion, 2009. FUSION'09. 12th
  International Conference on}}. IEEE, \bibinfo{pages}{756--763}.
\newblock


\bibitem[\protect\citeauthoryear{Lee, Pei, Hajiaghayi, Pefkianakis, Lu, Yan,
  Ge, Yates, and Kosseifi}{Lee et~al\mbox{.}}{2012}]%
        {historical-avg}
\bibfield{author}{\bibinfo{person}{Suk-Bok Lee}, \bibinfo{person}{Dan Pei},
  \bibinfo{person}{MohammadTaghi Hajiaghayi}, \bibinfo{person}{Ioannis
  Pefkianakis}, \bibinfo{person}{Songwu Lu}, \bibinfo{person}{He Yan},
  \bibinfo{person}{Zihui Ge}, \bibinfo{person}{Jennifer Yates}, {and}
  \bibinfo{person}{Mario Kosseifi}.} \bibinfo{year}{2012}\natexlab{}.
\newblock \showarticletitle{Threshold compression for 3g scalable monitoring}.
  In \bibinfo{booktitle}{\emph{INFOCOM, 2012 Proceedings IEEE}}. IEEE,
  \bibinfo{pages}{1350--1358}.
\newblock


\bibitem[\protect\citeauthoryear{Liu, Zhao, Xu, Sun, Pei, Luo, Jing, and
  Feng}{Liu et~al\mbox{.}}{2015}]%
        {opprentice}
\bibfield{author}{\bibinfo{person}{Dapeng Liu}, \bibinfo{person}{Youjian Zhao},
  \bibinfo{person}{Haowen Xu}, \bibinfo{person}{Yongqian Sun},
  \bibinfo{person}{Dan Pei}, \bibinfo{person}{Jiao Luo},
  \bibinfo{person}{Xiaowei Jing}, {and} \bibinfo{person}{Mei Feng}.}
  \bibinfo{year}{2015}\natexlab{}.
\newblock \showarticletitle{Opprentice: Towards Practical and Automatic Anomaly
  Detection Through Machine Learning}. In \bibinfo{booktitle}{\emph{Proceedings
  of the 2015 ACM Conference on Internet Measurement Conference}}
  \emph{(\bibinfo{series}{IMC '15})}. \bibinfo{publisher}{ACM},
  \bibinfo{address}{New York, NY, USA}, \bibinfo{pages}{211--224}.
\newblock
\showISBNx{978-1-4503-3848-6}
\urldef\tempurl%
\url{https://doi.org/10.1145/2815675.2815679}
\showDOI{\tempurl}


\bibitem[\protect\citeauthoryear{Lu and Ghorbani}{Lu and Ghorbani}{2009}]%
        {wavelet}
\bibfield{author}{\bibinfo{person}{Wei Lu} {and} \bibinfo{person}{Ali~A
  Ghorbani}.} \bibinfo{year}{2009}\natexlab{}.
\newblock \showarticletitle{Network anomaly detection based on wavelet
  analysis}.
\newblock \bibinfo{journal}{\emph{EURASIP Journal on Advances in Signal
  Processing}}  \bibinfo{volume}{2009} (\bibinfo{year}{2009}),
  \bibinfo{pages}{4}.
\newblock


\bibitem[\protect\citeauthoryear{Mahimkar, Ge, Wang, Yates, Zhang, Emmons,
  Huntley, and Stockert}{Mahimkar et~al\mbox{.}}{2011}]%
        {svd}
\bibfield{author}{\bibinfo{person}{Ajay Mahimkar}, \bibinfo{person}{Zihui Ge},
  \bibinfo{person}{Jia Wang}, \bibinfo{person}{Jennifer Yates},
  \bibinfo{person}{Yin Zhang}, \bibinfo{person}{Joanne Emmons},
  \bibinfo{person}{Brian Huntley}, {and} \bibinfo{person}{Mark Stockert}.}
  \bibinfo{year}{2011}\natexlab{}.
\newblock \showarticletitle{Rapid detection of maintenance induced changes in
  service performance}. In \bibinfo{booktitle}{\emph{Proceedings of the Seventh
  COnference on emerging Networking EXperiments and Technologies}}. ACM,
  \bibinfo{pages}{13}.
\newblock


\bibitem[\protect\citeauthoryear{M{\"u}nz, Li, and Carle}{M{\"u}nz
  et~al\mbox{.}}{2007}]%
        {k-means}
\bibfield{author}{\bibinfo{person}{Gerhard M{\"u}nz}, \bibinfo{person}{Sa Li},
  {and} \bibinfo{person}{Georg Carle}.} \bibinfo{year}{2007}\natexlab{}.
\newblock \showarticletitle{Traffic anomaly detection using k-means
  clustering}. In \bibinfo{booktitle}{\emph{GI/ITG Workshop MMBnet}}.
\newblock


\bibitem[\protect\citeauthoryear{Nicolau, McDermott, et~al\mbox{.}}{Nicolau
  et~al\mbox{.}}{2016}]%
        {kde}
\bibfield{author}{\bibinfo{person}{Miguel Nicolau}, \bibinfo{person}{James
  McDermott}, {et~al\mbox{.}}} \bibinfo{year}{2016}\natexlab{}.
\newblock \showarticletitle{One-Class Classification for Anomaly Detection with
  Kernel Density Estimation and Genetic Programming}. In
  \bibinfo{booktitle}{\emph{European Conference on Genetic Programming}}.
  Springer, \bibinfo{pages}{3--18}.
\newblock


\bibitem[\protect\citeauthoryear{Nielsen and Jensen}{Nielsen and
  Jensen}{2009}]%
        {graph}
\bibfield{author}{\bibinfo{person}{Thomas~Dyhre Nielsen} {and}
  \bibinfo{person}{Finn~Verner Jensen}.} \bibinfo{year}{2009}\natexlab{}.
\newblock \bibinfo{booktitle}{\emph{Bayesian networks and decision graphs}}.
\newblock \bibinfo{publisher}{Springer Science \& Business Media}.
\newblock


\bibitem[\protect\citeauthoryear{Pincombe}{Pincombe}{2005}]%
        {arma}
\bibfield{author}{\bibinfo{person}{Brandon Pincombe}.}
  \bibinfo{year}{2005}\natexlab{}.
\newblock \showarticletitle{Anomaly detection in time series of graphs using
  arma processes}.
\newblock \bibinfo{journal}{\emph{Asor Bulletin}} \bibinfo{volume}{24},
  \bibinfo{number}{4} (\bibinfo{year}{2005}), \bibinfo{pages}{2}.
\newblock


\bibitem[\protect\citeauthoryear{Rezende, Mohamed, and Wierstra}{Rezende
  et~al\mbox{.}}{2014}]%
        {rezende_stochastic_2014}
\bibfield{author}{\bibinfo{person}{Danilo~Jimenez Rezende},
  \bibinfo{person}{Shakir Mohamed}, {and} \bibinfo{person}{Daan Wierstra}.}
  \bibinfo{year}{2014}\natexlab{}.
\newblock \showarticletitle{Stochastic {Backpropagation} and {Approximate}
  {Inference} in {Deep} {Generative} {Models}}. In
  \bibinfo{booktitle}{\emph{Proceedings of the 31st {International}
  {Conference} on {International} {Conference} on {Machine} {Learning} -
  {Volume} 32}} \emph{(\bibinfo{series}{{ICML}'14})}.
  \bibinfo{publisher}{JMLR.org}, \bibinfo{address}{Beijing, China},
  \bibinfo{pages}{II--1278--II--1286}.
\newblock


\bibitem[\protect\citeauthoryear{Ringberg, Soule, Rexford, and Diot}{Ringberg
  et~al\mbox{.}}{2007}]%
        {sensitivity_pca}
\bibfield{author}{\bibinfo{person}{Haakon Ringberg}, \bibinfo{person}{Augustin
  Soule}, \bibinfo{person}{Jennifer Rexford}, {and} \bibinfo{person}{Christophe
  Diot}.} \bibinfo{year}{2007}\natexlab{}.
\newblock \showarticletitle{Sensitivity of PCA for Traffic Anomaly Detection}.
  In \bibinfo{booktitle}{\emph{Proceedings of the 2007 ACM SIGMETRICS
  International Conference on Measurement and Modeling of Computer Systems}}
  \emph{(\bibinfo{series}{SIGMETRICS '07})}. \bibinfo{publisher}{ACM},
  \bibinfo{address}{New York, NY, USA}, \bibinfo{pages}{109--120}.
\newblock
\showISBNx{978-1-59593-639-4}
\urldef\tempurl%
\url{https://doi.org/10.1145/1254882.1254895}
\showDOI{\tempurl}


\bibitem[\protect\citeauthoryear{Sejnowski and Rosenberg}{Sejnowski and
  Rosenberg}{1987}]%
        {sejnowski1987parallel}
\bibfield{author}{\bibinfo{person}{Terrence~J Sejnowski} {and}
  \bibinfo{person}{Charles~R Rosenberg}.} \bibinfo{year}{1987}\natexlab{}.
\newblock \showarticletitle{Parallel networks that learn to pronounce English
  text}.
\newblock \bibinfo{journal}{\emph{Complex systems}} \bibinfo{volume}{1},
  \bibinfo{number}{1} (\bibinfo{year}{1987}), \bibinfo{pages}{145--168}.
\newblock


\bibitem[\protect\citeauthoryear{Shanbhag and Wolf}{Shanbhag and Wolf}{2009}]%
        {normalization_schema}
\bibfield{author}{\bibinfo{person}{Shashank Shanbhag} {and}
  \bibinfo{person}{Tilman Wolf}.} \bibinfo{year}{2009}\natexlab{}.
\newblock \showarticletitle{Accurate anomaly detection through parallelism}.
\newblock \bibinfo{journal}{\emph{Network, IEEE}} \bibinfo{volume}{23},
  \bibinfo{number}{1} (\bibinfo{year}{2009}), \bibinfo{pages}{22--28}.
\newblock


\bibitem[\protect\citeauthoryear{S{\"o}lch, Bayer, Ludersdorfer, and van~der
  Smagt}{S{\"o}lch et~al\mbox{.}}{2016}]%
        {vi-storn}
\bibfield{author}{\bibinfo{person}{Maximilian S{\"o}lch},
  \bibinfo{person}{Justin Bayer}, \bibinfo{person}{Marvin Ludersdorfer}, {and}
  \bibinfo{person}{Patrick van~der Smagt}.} \bibinfo{year}{2016}\natexlab{}.
\newblock \showarticletitle{Variational inference for on-line anomaly detection
  in high-dimensional time series}.
\newblock \bibinfo{journal}{\emph{International Conference on Machine Laerning
  Anomaly detection Workshop}} (\bibinfo{year}{2016}).
\newblock


\bibitem[\protect\citeauthoryear{Sterne, White, Carlin, Spratt, Royston,
  Kenward, Wood, and Carpenter}{Sterne et~al\mbox{.}}{2009}]%
        {missing}
\bibfield{author}{\bibinfo{person}{Jonathan~AC Sterne}, \bibinfo{person}{Ian~R
  White}, \bibinfo{person}{John~B Carlin}, \bibinfo{person}{Michael Spratt},
  \bibinfo{person}{Patrick Royston}, \bibinfo{person}{Michael~G Kenward},
  \bibinfo{person}{Angela~M Wood}, {and} \bibinfo{person}{James~R Carpenter}.}
  \bibinfo{year}{2009}\natexlab{}.
\newblock \showarticletitle{Multiple imputation for missing data in
  epidemiological and clinical research: potential and pitfalls}.
\newblock \bibinfo{journal}{\emph{Bmj}}  \bibinfo{volume}{338}
  (\bibinfo{year}{2009}), \bibinfo{pages}{b2393}.
\newblock


\bibitem[\protect\citeauthoryear{Sutskever, Vinyals, and Le}{Sutskever
  et~al\mbox{.}}{2014}]%
        {RNN}
\bibfield{author}{\bibinfo{person}{Ilya Sutskever}, \bibinfo{person}{Oriol
  Vinyals}, {and} \bibinfo{person}{Quoc~V Le}.}
  \bibinfo{year}{2014}\natexlab{}.
\newblock \showarticletitle{Sequence to sequence learning with neural
  networks}. In \bibinfo{booktitle}{\emph{Advances in neural information
  processing systems}}. \bibinfo{pages}{3104--3112}.
\newblock


\bibitem[\protect\citeauthoryear{Wang and Yeung}{Wang and Yeung}{2016}]%
        {deep-bayes}
\bibfield{author}{\bibinfo{person}{Hao Wang} {and} \bibinfo{person}{Dit-Yan
  Yeung}.} \bibinfo{year}{2016}\natexlab{}.
\newblock \showarticletitle{Towards Bayesian deep learning: A survey}.
\newblock \bibinfo{journal}{\emph{arXiv preprint arXiv:1604.01662}}
  (\bibinfo{year}{2016}).
\newblock


\bibitem[\protect\citeauthoryear{Yaacob, Tan, Chien, and Tan}{Yaacob
  et~al\mbox{.}}{2010}]%
        {arima}
\bibfield{author}{\bibinfo{person}{Asrul~H Yaacob}, \bibinfo{person}{Ian~KT
  Tan}, \bibinfo{person}{Su~Fong Chien}, {and} \bibinfo{person}{Hon~Khi Tan}.}
  \bibinfo{year}{2010}\natexlab{}.
\newblock \showarticletitle{Arima based network anomaly detection}. In
  \bibinfo{booktitle}{\emph{Communication Software and Networks, 2010.
  ICCSN'10. Second International Conference on}}. IEEE,
  \bibinfo{pages}{205--209}.
\newblock


\bibitem[\protect\citeauthoryear{Yan, Flavel, Ge, Gerber, Massey, Papadopoulos,
  Shah, and Yates}{Yan et~al\mbox{.}}{2012}]%
        {holt-winters}
\bibfield{author}{\bibinfo{person}{He Yan}, \bibinfo{person}{Ashley Flavel},
  \bibinfo{person}{Zihui Ge}, \bibinfo{person}{Alexandre Gerber},
  \bibinfo{person}{Dan Massey}, \bibinfo{person}{Christos Papadopoulos},
  \bibinfo{person}{Hiren Shah}, {and} \bibinfo{person}{Jennifer Yates}.}
  \bibinfo{year}{2012}\natexlab{}.
\newblock \showarticletitle{Argus: End-to-end service anomaly detection and
  localization from an ISP's point of view}. In
  \bibinfo{booktitle}{\emph{INFOCOM, 2012 Proceedings IEEE}}. IEEE,
  \bibinfo{pages}{2756--2760}.
\newblock


\end{thebibliography}

\end{document}